\documentclass[conference]{IEEEtran}
\IEEEoverridecommandlockouts

\usepackage{cite}
\usepackage{amsmath,amssymb,amsfonts}
\usepackage{graphicx}
\usepackage{textcomp}
\usepackage{xcolor,colortbl}
\def\BibTeX{{\rm B\kern-.05em{\sc i\kern-.025em b}\kern-.08em
    T\kern-.1667em\lower.7ex\hbox{E}\kern-.125emX}}

\usepackage{circuitikz}
\usetikzlibrary{shapes.geometric, patterns, arrows.meta, shapes.arrows, fit, shadows.blur, decorations.pathreplacing}
\usepackage{pgfplots,pgfplotstable}
\usepackage{subfig}
\usepackage{threeparttable}
\usepackage{multirow,hhline}
\usepackage{xcolor, colortbl}
\usepackage{todonotes}
\usepackage{setspace}

\usepackage{collcell}
\usepackage{enumitem}
\usepackage{tikz}
\usetikzlibrary{matrix}
\usepackage{multirow}
\usepackage{booktabs}

\definecolor{babyblue}{rgb}{0.54,0.81,0.94}
\definecolor{mypink}{rgb}{0.858, 0.188, 0.478}

\definecolor{myyellow}{rgb}{1,0.965,0.890}
\definecolor{mypurple}{rgb}{0.804,0.757,1}
\definecolor{vsblue}{rgb}{0, 0, 0.706}
\definecolor{vspurple}{rgb}{0.5, 0, 0.5}
\definecolor{vsgray}{rgb}{0.5, 0.5, 0.5}
\definecolor{vsorange}{rgb}{0.773, 0.525, 0.0}

\definecolor{new_red}{rgb}{0.95,0.47,0.47}
\definecolor{new_green}{rgb}{0.85,0.98,0.77}

\usetikzlibrary{decorations.text}

\usepackage{enumitem}

\usepackage{dblfloatfix}

\usepgfplotslibrary{groupplots}

\usepackage{tikz}
\usetikzlibrary{arrows.meta, positioning, calc, shapes.geometric, fit}
\definecolor{ppred}{RGB}{183,30,66}
\definecolor{pppurple}{RGB}{121,95,175}

\usepackage{multirow}
\usepackage{makecell} 

\usepackage{algorithm}
\usepackage{algpseudocode}
\algrenewcommand{\algorithmiccomment}[1]{%
  \hfill\parbox[t]{0.56\linewidth}{\raggedright$\triangleright$ #1}}
\usepackage{amssymb}

\usepackage[hidelinks]{hyperref}

\begin{document}

\title{Inference Latency-Aware Tsetlin Machine Training\\
}

\author{
  \IEEEauthorblockN{Yefan Zeng, Shengyu Duan, Rishad Shafik, Alex Yakovlev}
  \IEEEauthorblockA{
    \textit{Microsystems Research Group, Newcastle University} \\
    Newcastle upon Tyne, United Kingdom \\
    \{y.zeng15, shengyu.duan, rishad.shafik, alex.yakovlev\}@newcastle.ac.uk
  }
}

\maketitle

\begin{abstract}
The Tsetlin Machine (TM) is a logic-based machine learning model well suited to low-latency inference on resource-constrained edge devices. Because each clause is an AND over Boolean literals, both the literals and their associated Tsetlin Automaton (TA) actions can be represented at the bit level, enabling bitwise evaluation and an early exit that is triggered as soon as any included literal evaluates to zero, forcing the clause output to zero. In this work, we study how the two TM training hyperparameters, $T$ and $s$, govern both classification accuracy and inference latency. Building on the inversely proportional $T–s$ relationship along which accuracy is nearly constant, we find that inference latency is reduced primarily by increasing $s$; rather than exhaustively searching the $(T, s)$ grid at high training cost, we propose a directed search strategy that treats accuracy as a constraint and advances toward larger $s$ to reduce inference latency within a fixed evaluation budget. Experimental results on the gem5 simulator with an ARM processor show that, for certain datasets, the search is able to increase $s$ while preserving the baseline accuracy, thereby reducing inference latency at negligible accuracy cost: across six datasets, 
the method reduces inference latency by up to 16.34\%, while some datasets exhibit an accuracy reduction of only 0.06\%.
The proposed strategy requires far fewer configuration evaluations than an exhaustive grid search.
\end{abstract}


\section{Introduction} \label{sec:intro}

Edge devices impose tight latency, memory and power budgets on machine
learning (ML) inference, which deep neural networks (DNNs) meet poorly
because their predictions rest on large numbers of multiply--accumulate
operations~\cite{li2023comprehensive}. Logic-based ML algorithms avoid this arithmetic
bottleneck by encoding learned patterns in Boolean form and evaluating them
with bitwise logic. 
The Tsetlin Machine (TM) is a logic-based ML algorithm that employs Tsetlin Automata (TAs) as its fundamental learning units to construct multiple logical pattern recognizers, referred to as clauses.
A clause can be represented as a propositional conjunction (AND formula), making the TM inherently interpretable by construction rather than relying on post-hoc explanations.
In addition, the TM eliminates the need for floating-point arithmetic during both training and inference, making it computationally efficient and cost-effective.

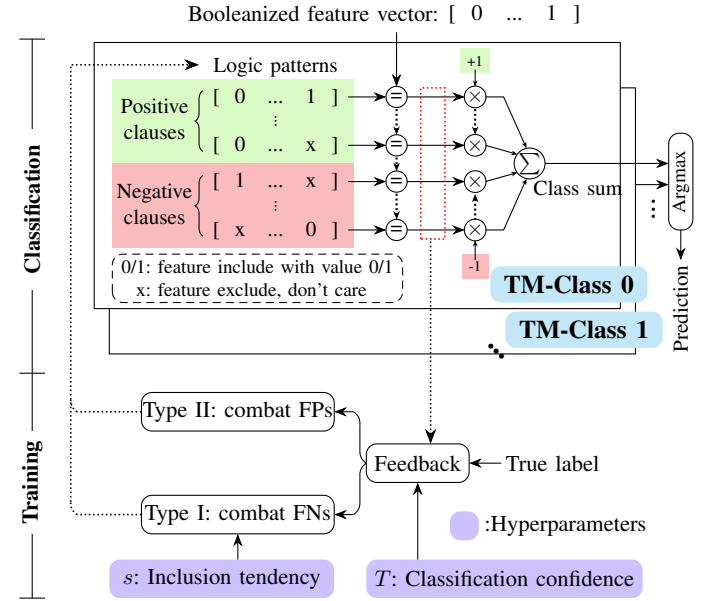
\begin{figure}[!htb]
	\centering
	\begin{tikzpicture}[every node/.style={scale=1}, scale=1]
	
	\begin{scope}[xshift=0.5cm, yshift=-3.8cm, every node/.style={scale=0.7}, scale=0.7]	
	\node[] at (0,0) {\large Booleanized feature vector:};
	\node[] at (3.8,0) {\large [ \ 0 \ \ ... \ \ 1 \ ]};
	\end{scope}
	
	\begin{scope}[xshift=0.2cm,yshift=-5.1cm,every node/.style={scale=0.8}, scale=0.8]
	\draw[fill=white] (-3,-4) rectangle ++(8.7,4.4);
	\draw[draw=none, fill=babyblue!50!white, rounded corners] (3.55,-3.9) rectangle ++(2.6,0.6) node[pos=.5] () {\large \textbf{TM-Class 1}};
	\node[rotate=-45] at (3.4,-3.95) {\Large{\textbf{...}}};
	\draw[-Stealth] (5.7, -1.2) --++ (0.55,0);
	\node[rotate=90] at (6,-1.6) () {\Large ...};	

	\end{scope}
	
	\begin{scope}[xshift=0cm,yshift=-4.5cm,every node/.style={scale=0.8}, scale=0.8]
	\draw[fill=white] (-3,-4) rectangle ++(8.7,4.4);
	
	\node[] at (0,0) {Logic patterns};
	
	\draw[draw=none, fill=new_green] (-2.7,-1.6) rectangle ++(4,1.4);
	\node[] at (0,-0.5) {[ \ 0 \ \ ... \ \ 1 \ ]};
	\node[rotate=90] at (0,-0.9) {\footnotesize ...};
	\node[] at (0,-1.3) {[ \ 0 \ \ ... \ \ x \ ]};
	\node [] at (-2,-0.9) {\begin{tabular}{@{}c@{}} Positive \\ clauses \end{tabular}};
	\draw[decorate,decoration={brace,amplitude=3pt,mirror}] (-1.2,-0.4) -- (-1.2,-1.4);
	\node[draw,circle,minimum size=0.35cm,inner sep=0pt,fill=white] at (2,-0.5) {=};
	\draw[-Stealth] (1.2,-0.5) -- (1.83,-0.5);
	\node[draw,circle,minimum size=0.35cm,inner sep=0pt,fill=white] at (2,-1.3) {=};
	\draw[-Stealth] (1.2,-1.3) -- (1.83,-1.3);
	\node[draw,circle,minimum size=0.35cm,inner sep=0pt,fill=white] at (3.3,-0.5) {$\times$};
	\draw[-Stealth] (2.17,-0.5) -- (3.13,-0.5);
	\node[draw,circle,minimum size=0.35cm,inner sep=0pt,fill=white] at (3.3,-1.3) {$\times$};
	\draw[-Stealth] (2.17,-1.3) -- (3.13,-1.3);
	\draw[thick, arrows={-Stealth[scale=0.5]}, densely dotted] (3.3,-0.7) -- (3.3,-1.13);
	\node[fill=new_green] at (3.3,0.1) {\scriptsize +1};
	\draw[arrows={-Stealth[scale=0.5]}] (3.3,-0.05) -- (3.3,-0.33);	   

	\draw[draw=none, fill=new_red!50!white] (-2.7,-1.6) rectangle ++(4,-1.4);	
	\node[] at (0,-1.9) {[ \ 1 \ \ ... \ \ x \ ]};
	\node[rotate=90] at (0,-2.3) {\footnotesize ...};
	\node[] at (0,-2.7) {[ \ x \ \ ... \ \ 0 \ ]};
	\node [] at (-2,-2.3) {\begin{tabular}{@{}c@{}} Negative \\ clauses \end{tabular}};
	\draw[decorate,decoration={brace,amplitude=3pt,mirror}] (-1.2,-1.8) -- (-1.2,-2.8);
	\node[draw,circle,minimum size=0.35cm,inner sep=0pt,fill=white] at (2,-1.9) {=};
	\draw[-Stealth] (1.2,-1.9) -- (1.83,-1.9);
	\node[draw,circle,minimum size=0.35cm,inner sep=0pt,fill=white] at (2,-2.7) {=};
	\draw[-Stealth] (1.2,-2.7) -- (1.83,-2.7);
	\node[draw,circle,minimum size=0.35cm,inner sep=0pt,fill=white] at (3.3,-1.9) {$\times$};
	\draw[-Stealth] (2.17,-1.9) -- (3.13,-1.9);
	\node[draw,circle,minimum size=0.35cm,inner sep=0pt,fill=white] at (3.3,-2.7) {$\times$};
	\draw[-Stealth] (2.17,-2.7) -- (3.13,-2.7); 
	\draw[thick, arrows={Stealth[scale=0.5]-}, densely dotted] (3.3,-2.1) -- (3.3,-2.53);
	\node[fill=new_red!50!white] at (3.3,-3.3) {\scriptsize -1};
	\draw[arrows={-Stealth[scale=0.5]}] (3.3,-3.2) -- (3.3,-2.87);	  
	
	\draw[thick, arrows={-Stealth[scale=0.5]}, densely dotted] (2,-0.7) -- (2,-1.13);
	\draw[thick, arrows={-Stealth[scale=0.5]}, densely dotted] (2,-1.5) -- (2,-1.73);
	\draw[thick, arrows={-Stealth[scale=0.5]}, densely dotted] (2,-2.1) -- (2,-2.53);     
	
	\node[] at (-0.3,-3.3) {\small 0/1: feature include with value 0/1};
	\node[] at (-0.4,-3.7) {\small x: feature exclude, don't care};
	\draw[rounded corners,densely dashed] (-2.7,-3.9) rectangle ++(4.8,0.8);
	
	\node[draw,circle,minimum size=0.5cm,inner sep=0pt,fill=white] at (4.2,-1.6) {$\sum$};
	\draw[arrows={-Stealth[scale=0.5]}] (3.47,-0.5) --++ (0.3,0) -- (4.15,-1.35);
	\draw[arrows={-Stealth[scale=0.5]}] (3.47,-2.7) --++ (0.3,0) -- (4.15,-1.85);
	\draw[arrows={-Stealth[scale=0.5]}] (3.47,-1.3) -- (4,-1.45);
	\draw[arrows={-Stealth[scale=0.5]}] (3.47,-1.9) -- (4,-1.75);
	
	\draw[draw=none, fill=babyblue!50!white, rounded corners] (3.55,-3.9) rectangle ++(2.6,0.6) node[pos=.5] () {\large \textbf{TM-Class 0}};	
	
	\draw[-Stealth] (4.45, -1.6) --++ (2.05,0);
	\node[] at (5, -2) {Class sum};	
	
	\draw[rounded corners] (6.5,-2.7) rectangle ++(0.4,1.6) node[pos=.5, rotate=90] () {\small Argmax};
	\draw[-Stealth] (6.7,-2.7) --++ (0,-0.5);
	\node[rotate=90] at (6.7,-4) {Prediction};
	
	\draw[Stealth-] (2,-0.33) --++ (0,0.9);
	
	\draw[red, fill=none, densely dotted, semithick] (2.4,-2.85) rectangle ++(0.4,2.5);
	   
	\end{scope}
	
	\begin{scope}[xshift=1.2cm, yshift=-10cm, every node/.style={scale=0.85}, scale=0.85]
	\draw[rounded corners] (0,0) rectangle ++(1.6,0.6) node[pos=.5] () {Feedback};
	\draw[Stealth-] (1.6,0.3) --++ (0.5,0);	
	\node[] at (2.9,0.3) {True label};
	\draw[Stealth-, densely dotted, semithick] (1,0.6) --++ (0,3.2); 
	\draw[draw=none, fill=mypurple, rounded corners] (0,-1.8) rectangle ++(4.3,0.6) node[pos=.5] () {$T$: Classification confidence};
	\draw[Stealth-] (0.8,0) --++ (0,-1.2); 
	
	\draw[rounded corners] (-3.5,-0.8) rectangle ++(3,0.6) node[pos=.5] () {Type I: combat FNs};
	\draw[] (0,0.3) to [bend right] (-0.1,0.1);
	\draw[] (-0.1,0.1) -- (-0.1,-0.3);
	\draw[] (-0.1,-0.3) to [bend left] (-0.2,-0.5);
	\draw[-Stealth] (-0.2,-0.5) -- (-0.5,-0.5);
	
	\draw[rounded corners] (-3.5,0.8) rectangle ++(3,0.6) node[pos=.5] () {Type II: combat FPs};
	\draw[] (0,0.3) to [bend left] (-0.1,0.5);
	\draw[] (-0.1,0.5) -- (-0.1,0.9);
	\draw[] (-0.1,0.9) to [bend right] (-0.2,1.1);
	\draw[-Stealth] (-0.2,1.1) -- (-0.5,1.1);
	
	\draw[draw=none, fill=mypurple, rounded corners] (-4,-1.8) rectangle ++(3.5,0.6) node[pos=.5] () {$s$: Inclusion tendency};
	\draw[Stealth-] (-2,-0.8) --++ (0,-0.4);
	
	\draw[densely dotted, semithick] (-3.5,-0.5) -- (-4.5,-0.5);
	\draw[densely dotted, semithick] (-4.5,-0.5) to [bend left] (-4.6,-0.3);
	\draw[densely dotted, semithick] (-3.5,1.1) -- (-4.5,1.1);
	\draw[densely dotted, semithick] (-4.5,1.1) to [bend left] (-4.6,1.3);	
	\draw[-Stealth, densely dotted, semithick] (-4.6,-0.3) --++ (0,6.8) --++ (2,0);	
	
	\draw[draw=none, fill=mypurple, rounded corners] (1.3,-0.9) rectangle ++(0.45,0.45) node[pos=.5] () {};
	\node[] at (3.1,-0.675) {:Hyperparameters};	
	
	\draw[] (-5.2,-1.8) -- (-5.2,6.9);
	\draw[]  (-5.4,-1.8) -- (-5,-1.8);
	\draw[]  (-5.4,6.9) -- (-5,6.9);
	\draw[]	 (-5.4,1.7) -- (-5,1.7);
	
	\draw[fill=white,draw=none] (-5.5,-0.8) rectangle ++(0.6,1.6) node[pos=.5, rotate=90] () {\textbf{Training}};
	\draw[fill=white,draw=none] (-5.5,3) rectangle ++(0.6,2.5) node[pos=.5, rotate=90] () {\textbf{Classification}};
	
	\end{scope}
	
	\end{tikzpicture}
	\caption{TM classification architecture and training. Type I/II feedback combats false negatives (FNs)/false positives (FPs).
	}
	\label{fig:tm_architecture}
\end{figure}

Fig.~\ref{fig:tm_architecture} illustrates the TM classification architecture and provides an overview of its training mechanism.
Specifically, a TM operates exclusively on Boolean data. Therefore, raw features must first be transformed into Boolean features, typically through thresholding.
For classification, a TM model consists of one TM for each class, with each TM learning the Boolean patterns of its corresponding class. 
Within a TM, a clause is activated (output=1) only when every feature in its learned pattern matches the input; a single mismatched feature is sufficient to deactivate the clause (output=0). Activated positive and negative clauses cast votes of $+1$ and $-1$, respectively, which are accumulated into a class score. The predicted class is the one with the highest score.

As can be seen from Fig.~\ref{fig:tm_architecture}, not all features are included in a clause. Feature inclusion and exclusion are determined by TAs through a reinforcement feedback mechanism that compares the clause outputs with the ground-truth labels. A hyperparameter, $T$, serves as the classification confidence (voting target) by encouraging the class sum to approach $\pm T$; the farther the class sum deviates from these targets, the more likely feedback is to be triggered. Two types of feedback, Type I and Type II, are employed to reduce false negatives (FNs) and false positives (FPs), respectively. Another hyperparameter, $s$, controls the specificity of the learned clauses by regulating the tendency to include features, in Type I feedback. Larger values of $s$ encourage the inclusion of more features in each clause. A more detailed description of the training mechanism can be found in previous publications \cite{Granmo2018TM, duan2025ethereal}.

The conventional approach to selecting the optimal values of $T$ and $s$ is to perform an exhaustive grid search over candidate parameter combinations and retain the combination that achieves the highest classification accuracy \cite{gronningsaeter2024optimized, thangaraj2025interpretable, abbas2025ai}. 
Every candidate demands a full training run, so the cost grows
with the resolution of the grid and the workflow becomes time-consuming. A more
directed alternative has been made possible by an existing characterization
of the $(T, s)$ space: the generally inversely proportional relationship between $T$
and $s$ for the optimal accuracy has been reported in a previous paper,
which derives a mathematical model of the two hyperparameters and evaluates
it on MNIST~\cite{tarasyuk2023systematic}. 
What both the exhaustive search and this mathematical model share, however, is
that they treat $(T, s)$ selection as a question of accuracy alone.

Our observation is that $T$ and $s$ also govern inference latency. As
established above, they fix the clause expression, and in particular the
number of literals each clause includes. This determines how many clauses
evaluate to zero on a given input, and 
the position at which a mismatch is encountered during clause evaluation directly determines how frequently inference can terminate clause evaluation early~\cite{zeng2025fast}.
An
accuracy-optimal $(T, s)$ therefore carries no guarantee of being a
latency-efficient one, and to the best of our knowledge, 
the relationship between the hyperparameters $T$ and $s$ and the resulting trade-off between accuracy and inference latency remains largely unexplored.

This paper studies accuracy-latency trade-off of TM, based on different values of $T$ and $s$. We first evaluate the impact of $T$ and
$s$ on accuracy and inference latency 
and find
these two objectives are largely decoupled. 
Specifically, inference latency is largely insensitive to movement along the
inversely proportional relationship that preserves accuracy, and falls
instead with increasing $s$. 
Based on this observation, 
we propose an
inference latency-aware training strategy. Rather than sampling the design
space indiscriminately, it advances towards larger $s$ to reduce latency,
compensates with $T$ to hold accuracy, and admits a configuration only if a
user-specified accuracy constraint is still met. 
A fixed budget on training evaluations bounds the search process, ensuring a linear number of evaluations with respect to the explored $(T, s)$ configurations.

We validate the strategy on six well-established benchmark datasets
with inference latency measured 
on an ARM processor modeled in gem5~\cite{10.1145/2024716.2024718}.
The results demonstrate that the proposed strategy reduces inference latency by 0.25\%–16.34\% while incurring only a 0.06\%–1.97\% reduction in accuracy. Furthermore, when searching for a model that satisfies specific accuracy and latency requirements, the proposed strategy ensures that the number of evaluations over different $(T, s)$ combinations increases linearly from an initial accuracy–latency configuration.


Our contributions are as follows:
\begin{itemize}
  \item 
        A systematic exploration of the $(T, s)$ design space reveals that accuracy and inference latency are largely decoupled, enabling latency optimization with minimal impact on accuracy.

  \item 
  An efficient training strategy that substantially lowers the evaluation cost of  $(T, s)$ design space exploration, compared with exhaustive search.

  \item 
Comprehensive end-to-end validation of Booleanization, TM training, hyperparameter tuning and deployment on well-established benchmark datasets, demonstrating practical applicability across diverse tasks.

\end{itemize}

\section{TM Software Implementation} \label{sec:software}

Existing software implementations of TM inference fall into two classes, distinguished by what the model representation retains. 
The first retains the complete information---both includes and excludes.
This keeps the model aligned with the input, so that inference can be performed per sample, which makes it well suited to real-time computing.
The second retains only part of the information, namely the includes, and compresses the model by discarding the excludes through an
include-encoding scheme. Its parallel computing, however, is performed at batch level, which makes it better suited to a dataset that is
already available in full---such as a tabular dataset---rather than one whose samples arrive one at a time.

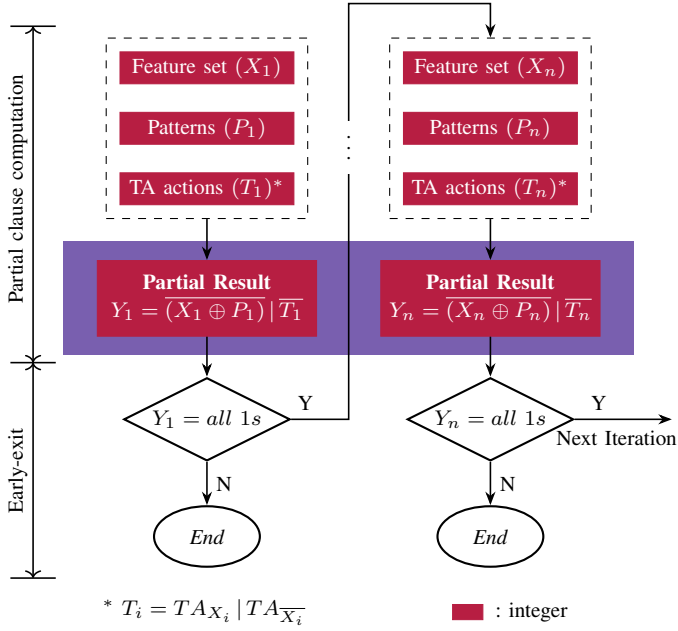
\begin{figure}[t]
\centering
\begin{tikzpicture}[
    box/.style={rectangle, draw=ppred, fill=ppred, text=white, align=center,
                font=\fontsize{8}{9.6}\selectfont, inner sep=2pt},
    dec/.style={diamond, draw=black, thick, align=center,
                font=\fontsize{8}{9.6}\selectfont, inner sep=1pt},
    term/.style={ellipse, draw=black, thick, align=center,
                 font=\fontsize{8}{9.6}\selectfont\itshape,
                 minimum width=1.4cm, minimum height=0.8cm},
    grp/.style={rectangle, draw=black, dashed, thin, inner sep=5pt},
    lbl/.style={font=\fontsize{8}{9.6}\selectfont, align=center},
    arr/.style={-{Stealth[length=1.8mm]}, draw=black, semithick},
]

\def\xL{0}      
\def\xR{3.75}    
\pgfmathsetmacro{\xM}{(\xL+\xR)/2}  

\node[box, minimum width=2.3cm] (fL) at (\xL,0)      {Feature set $(X_1)$};
\node[box, minimum width=2.3cm] (pL) at (\xL,-0.8)  {Patterns $(P_1)$};
\node[box, minimum width=2.3cm] (tL) at (\xL,-1.6)  {TA actions $(T_1)^*$};
\node[grp, fit=(fL)(pL)(tL)] (grpL) {};

\node[box, minimum width=2.3cm] (fR) at (\xR,0)      {Feature set $(X_n)$};
\node[box, minimum width=2.3cm] (pR) at (\xR,-0.8)  {Patterns $(P_n)$};
\node[box, minimum width=2.3cm] (tR) at (\xR,-1.6)  {TA actions $(T_n)^*$};
\node[grp, fit=(fR)(pR)(tR)] (grpR) {};


\fill[pppurple] (\xL-1.9,-3.8) rectangle (\xR+1.9,-2.3); 
\node[box, minimum width=2.9cm, minimum height=1.0cm] (rL) at (\xL,-3.05)
    {\textbf{Partial Result}\\[1pt]$Y_1=\overline{(X_1\oplus P_1)}\,|\,\overline{T_1}$};
\node[box, minimum width=2.9cm, minimum height=1.0cm] (rR) at (\xR,-3.05)
    {\textbf{Partial Result}\\[1pt]$Y_n=\overline{(X_n\oplus P_n)}\,|\,\overline{T_n}$};
\draw[arr] (grpL.south) -- (rL);
\draw[arr] (grpR.south) -- (rR);

\node[dec, minimum width=2.2cm, minimum height=1.1cm] (dL) at (\xL,-4.65) {};
\node[lbl] at (\xL,-4.65) {$Y_1=all\ 1s$};

\node[dec, minimum width=2.2cm, minimum height=1.1cm] (dR) at (\xR,-4.65) {};
\node[lbl] at (\xR,-4.65) {$Y_n=all\ 1s$};

\draw[arr] (rL.south) -- (dL.north);
\draw[arr] (rR.south) -- (dR.north);

\node[term] (eL) at (\xL,-6.2) {End};
\node[term] (eR) at (\xR,-6.2) {End};
\draw[arr] (dL.south) -- (eL.north) node[midway, right, lbl] {N};
\draw[arr] (dR.south) -- (eR.north) node[midway, right, lbl] {N};


\draw[arr] (dL.east) -- (\xM,-4.65) node[near start, above, lbl] {Y}
      -- (\xM,0.85) -| (grpR.north);
\fill[white] (\xM-0.13,-1.40) rectangle (\xM+0.13,-0.70);
\node[lbl] at (\xM,-0.95) {$\vdots$};

\draw[arr] (dR.east) -- (\xR+2.4,-4.65) node[near start, above, lbl] {Y};
\node[lbl, anchor=west] at (\xR+0.75,-4.9) {Next Iteration};

\draw[arr,<->] (\xL-2.3,0.55) -- (\xL-2.3,-3.9);
\node[lbl, rotate=90] at (\xL-2.5,-1.6) {Partial clause computation};
\draw[arr,<->] (\xL-2.3,-3.9) -- (\xL-2.3,-6.75);
\node[lbl, rotate=90] at (\xL-2.5,-5.325) {Early-exit};

\draw[black, semithick] (\xL-2.6,0.55)  -- (\xL-2,0.55);   
\draw[black, semithick] (\xL-2.6,-3.9) -- (\xL-2,-3.9);  
\draw[black, semithick] (\xL-2.6,-6.75) -- (\xL-2,-6.75);  

\node[lbl, anchor=west] at (\xL-1.5,-7.2)
    {$^{*}$ $ T_i=TA_{X_i}\,|\,TA_{\overline{X_i}}$};
\node[box, minimum width=0.4cm, minimum height=0.2cm] (leg) at (\xR-0.3,-7.2) {};
\node[lbl, anchor=west, right=1pt of leg] {: integer};

\end{tikzpicture}
\caption{Software implementation of TM inference with early exit mechanism.}
\label{fig:early_exit}
\end{figure}

The implementation of~\cite{zeng2025fast} belongs to the first class. It retains the action of every TA and casts both the model and the input into a
form on which clause evaluation reduces to bitwise operations, so that no include is addressed explicitly and no index is decoded at runtime. 
This work builds on the same early exit mechanism, which Fig.~\ref{fig:early_exit} summarises. The Boolean features of a datapoint are bit-packed into $n$ integers, where $n$ is the number of features divided by the word width of the target device---32 or 64 on typical edge processors, and wider where the device supports it. The trailing integer is zero-padded, which is safe because padded positions default to exclude. Each feature word $X_i$ is evaluated against two words derived from the same TA actions: $T_i = TA_{X_i} \,|\, TA_{\overline{X_i}}$, the OR of the positive and
negated halves of the model, which marks the positions at which a literal is included; and $P_i$, the sub-pattern the clause expects, which records the feature value each included literal requires. The partial result of the $i$-th word is $Y_i=\overline{(X_i \oplus P_i)}\,|\,\overline{T_i}$, so that a bit of $Y_i$ is~0 exactly where an included position is contradicted by the feature, excluded positions being masked to~1 by $\overline{T_i}$. Each iteration evaluates one word and decides on it. If $Y_i$ is all-ones the clause remains satisfiable and the next iteration proceeds; otherwise a mismatch on an included literal has been found, the clause output is determined to be~0, and the remaining $n-i$ words are never touched---an early exit. Because a word holds the features of a single datapoint, each input is classified independently, and the measured latency corresponds directly to the response time of one sample.

%

REDRESS~\cite{maheshwari2023redress} belongs to the second class. Since trained TA are
extremely sparse and only the includes contribute to the clause output,
its include-encoding scheme stores each include as a 16-bit integer
carrying the literal offset and stores nothing for the excludes, which
losslessly compresses the model by two orders of magnitude. Parallel
computing is then organised at batch level: rather than packing the
features of one sample into a machine word, REDRESS packs the same
feature of $W$ samples, so that a single pass over the model classifies
$W$ datapoints at once. This requires the samples to be collected and
bit-packed before inference begins, which is only possible when the
dataset is already available in full. Latency is consequently amortised
over a batch rather than attributable to a single input, and no
per-sample early exit is available.

In this paper, we use the first software implementation to evaluate inference latency. However, we observe that the impact of hyperparameters on latency and the proposed training strategy can also serve as guidelines for REDRESS optimization. A detailed discussion is provided in Section~\ref{sec:conc}.

\section{Inference-Latency Aware Training}

\subsection{Impact of $T$ and $s$ on Accuracy}  \label{subsec:ts_accuracy}

The key hyperparameters $T$ and $s$ govern the behavior of the TM training process and consequently have a significant impact on the overall classification accuracy \cite{tarasyuk2023systematic}. To investigate their effects, we train TM models with 300 clauses per class on the MNIST dataset using a wide range of $T$ and $s$ combinations. The resulting test accuracies are presented in Fig.~\ref{fig:Ts_accuracy}, where local optima are highlighted. As shown, the optimal values of $T$ and $s$ exhibit an approximately inverse relationship: similar peak accuracies can be achieved by increasing $T$ while decreasing $s$, or vice versa. This observation is consistent with the trend reported in \cite{tarasyuk2023systematic}.

\pgfplotsset{
	colormap={my colormap}{
rgb=(0.24220000,0.15040000,0.66030000)
rgb=(0.25039048,0.16499524,0.70761429)
rgb=(0.25777143,0.18178095,0.75113810)
rgb=(0.26472857,0.19775714,0.79521429)
rgb=(0.27064762,0.21467619,0.83637143)
rgb=(0.27511429,0.23423810,0.87098571)
rgb=(0.27830000,0.25587143,0.89907143)
rgb=(0.28033333,0.27823333,0.92210000)
rgb=(0.28133810,0.30059524,0.94137619)
rgb=(0.28101429,0.32275714,0.95788571)
rgb=(0.27946667,0.34467143,0.97167619)
rgb=(0.27597143,0.36668095,0.98290476)
rgb=(0.26991429,0.38920000,0.99060000)
rgb=(0.26024286,0.41232857,0.99515714)
rgb=(0.24403333,0.43583333,0.99883333)
rgb=(0.22064286,0.46025714,0.99728571)
rgb=(0.19633333,0.48471905,0.98915238)
rgb=(0.18340476,0.50737143,0.97979524)
rgb=(0.17864286,0.52885714,0.96815714)
rgb=(0.17643810,0.54990476,0.95201905)
rgb=(0.16874286,0.57026190,0.93587143)
rgb=(0.15400000,0.59020000,0.92180000)
rgb=(0.14602857,0.60911905,0.90785714)
rgb=(0.13802381,0.62762857,0.89729048)
rgb=(0.12481429,0.64592857,0.88834286)
rgb=(0.11125238,0.66350000,0.87631429)
rgb=(0.09520952,0.67982857,0.85978095)
rgb=(0.06887143,0.69477143,0.83935714)
rgb=(0.02966667,0.70816667,0.81633333)
rgb=(0.00357143,0.72026667,0.79170000)
rgb=(0.00665714,0.73121429,0.76601429)
rgb=(0.04332857,0.74109524,0.73940952)
rgb=(0.09639524,0.75000000,0.71203810)
rgb=(0.14077143,0.75840000,0.68415714)
rgb=(0.17170000,0.76696190,0.65544286)
rgb=(0.19376667,0.77576667,0.62510000)
rgb=(0.21608571,0.78430000,0.59230000)
rgb=(0.24695714,0.79179524,0.55674286)
rgb=(0.29061429,0.79729048,0.51882857)
rgb=(0.34064286,0.80080000,0.47885714)
rgb=(0.39090000,0.80287143,0.43544762)
rgb=(0.44562857,0.80241905,0.39091905)
rgb=(0.50440000,0.79930000,0.34800000)
rgb=(0.56156190,0.79423333,0.30448095)
rgb=(0.61739524,0.78761905,0.26123810)
rgb=(0.67198571,0.77927143,0.22270000)
rgb=(0.72420000,0.76984286,0.19102857)
rgb=(0.77383333,0.75980476,0.16460952)
rgb=(0.82031429,0.74981429,0.15352857)
rgb=(0.86343333,0.74060000,0.15963333)
rgb=(0.90354286,0.73302857,0.17741429)
rgb=(0.93925714,0.72878571,0.20995714)
rgb=(0.97275714,0.72977143,0.23944286)
rgb=(0.99564762,0.74337143,0.23714762)
rgb=(0.99698571,0.76585714,0.21994286)
rgb=(0.99520476,0.78925238,0.20276190)
rgb=(0.98920000,0.81356667,0.18853333)
rgb=(0.97862857,0.83862857,0.17655714)
rgb=(0.96764762,0.86390000,0.16429048)
rgb=(0.96100952,0.88901905,0.15367619)
rgb=(0.95967143,0.91345714,0.14225714)
rgb=(0.96279524,0.93733810,0.12650952)
rgb=(0.96911429,0.96062857,0.10636190)
rgb=(0.97690000,0.98390000,0.08050000)
	}
}

\begin{figure}[!htb]
	\centering

\begin{tikzpicture}[font=\normalsize, scale=1]
    \begin{axis}[
        xlabel=$T$,
        ylabel=\large{$s$},
        ylabel style={rotate=-90},
        xtick={4,6,8,10,12,14,16,18,20},
        ytick={4,6,8,10,12,14,16,18,20},
        title=Best accuracy (\%) in 100 epochs,
        title style={yshift=-0.2cm},
        view={0}{90},
        colorbar,colormap access=map,
        colormap name = my colormap,
        colorbar/width=2.5mm,
        colorbar style={font=\small,xshift=0cm},
        clip=false
    ]
        \addplot3 [surf,mesh/cols=10] table {Shengyu/Data/Ts_accuracy.dat};
        
\node () at (axis cs:4,4,100) {\scriptsize{\textcolor{white}{93.75}}};
\node () at (axis cs:6,4,100) {\scriptsize{\textcolor{white}{95.67}}};
\node () at (axis cs:8,4,100) {\scriptsize{95.80}};
\node () at (axis cs:10,4,100) {\scriptsize{96.51}};
\node () at (axis cs:12,4,100) {\scriptsize{96.56}};
\node () at (axis cs:14,4,100) {\scriptsize{96.90}};
\node () at (axis cs:16,4,100) {\scriptsize{97.00}};
\node () at (axis cs:18,4,100) {\scriptsize{97.27}};
\node () at (axis cs:20,4,100) {\scriptsize{97.18}};
\node () at (axis cs:4,6,100) {\scriptsize{\textcolor{white}{95.40}}};
\node () at (axis cs:6,6,100) {\scriptsize{96.30}};
\node () at (axis cs:8,6,100) {\scriptsize{96.64}};
\node () at (axis cs:10,6,100) {\scriptsize{96.90}};
\node () at (axis cs:12,6,100) {\scriptsize{97.12}};
\node () at (axis cs:14,6,100) {\scriptsize{97.12}};
\node () at (axis cs:16,6,100) {\scriptsize{97.32}};
\node () at (axis cs:18,6,100) {\scriptsize{97.24}};
\node () at (axis cs:20,6,100) {\scriptsize{97.30}};
\node () at (axis cs:4,8,100) {\scriptsize{95.94}};
\node () at (axis cs:6,8,100) {\scriptsize{96.54}};
\node () at (axis cs:8,8,100) {\scriptsize{97.04}};
\node () at (axis cs:10,8,100) {\scriptsize{97.07}};
\node () at (axis cs:12,8,100) {\scriptsize{97.41}};
\node () at (axis cs:14,8,100) {\scriptsize{97.44}};
\node () at (axis cs:16,8,100) {\scriptsize{97.38}};
\node () at (axis cs:18,8,100) {\scriptsize{97.22}};
\node () at (axis cs:20,8,100) {\scriptsize{96.99}};
\node () at (axis cs:4,10,100) {\scriptsize{96.16}};
\node () at (axis cs:6,10,100) {\scriptsize{96.57}};
\node () at (axis cs:8,10,100) {\scriptsize{96.91}};
\node () at (axis cs:10,10,100) {\scriptsize{97.30}};
\node () at (axis cs:12,10,100) {\scriptsize{97.13}};
\node () at (axis cs:14,10,100) {\scriptsize{97.21}};
\node () at (axis cs:16,10,100) {\scriptsize{97.15}};
\node () at (axis cs:18,10,100) {\scriptsize{96.95}};
\node () at (axis cs:20,10,100) {\scriptsize{96.70}};
\node () at (axis cs:4,12,100) {\scriptsize{96.46}};
\node () at (axis cs:6,12,100) {\scriptsize{96.69}};
\node () at (axis cs:8,12,100) {\scriptsize{97.00}};
\node () at (axis cs:10,12,100) {\scriptsize{97.37}};
\node () at (axis cs:12,12,100) {\scriptsize{97.11}};
\node () at (axis cs:14,12,100) {\scriptsize{96.99}};
\node () at (axis cs:16,12,100) {\scriptsize{96.92}};
\node () at (axis cs:18,12,100) {\scriptsize{96.78}};
\node () at (axis cs:20,12,100) {\scriptsize{96.31}};
\node () at (axis cs:4,14,100) {\scriptsize{96.24}};
\node () at (axis cs:6,14,100) {\scriptsize{96.76}};
\node () at (axis cs:8,14,100) {\scriptsize{97.11}};
\node () at (axis cs:10,14,100) {\scriptsize{97.05}};
\node () at (axis cs:12,14,100) {\scriptsize{97.00}};
\node () at (axis cs:14,14,100) {\scriptsize{96.87}};
\node () at (axis cs:16,14,100) {\scriptsize{96.66}};
\node () at (axis cs:18,14,100) {\scriptsize{96.29}};
\node () at (axis cs:20,14,100) {\scriptsize{96.02}};
\node () at (axis cs:4,16,100) {\scriptsize{96.01}};
\node () at (axis cs:6,16,100) {\scriptsize{96.72}};
\node () at (axis cs:8,16,100) {\scriptsize{97.02}};
\node () at (axis cs:10,16,100) {\scriptsize{97.02}};
\node () at (axis cs:12,16,100) {\scriptsize{96.78}};
\node () at (axis cs:14,16,100) {\scriptsize{96.77}};
\node () at (axis cs:16,16,100) {\scriptsize{96.20}};
\node () at (axis cs:18,16,100) {\scriptsize{96.01}};
\node () at (axis cs:20,16,100) {\scriptsize{95.70}};
\node () at (axis cs:4,18,100) {\scriptsize{96.44}};
\node () at (axis cs:6,18,100) {\scriptsize{96.78}};
\node () at (axis cs:8,18,100) {\scriptsize{96.91}};
\node () at (axis cs:10,18,100) {\scriptsize{96.81}};
\node () at (axis cs:12,18,100) {\scriptsize{96.96}};
\node () at (axis cs:14,18,100) {\scriptsize{96.47}};
\node () at (axis cs:16,18,100) {\scriptsize{96.07}};
\node () at (axis cs:18,18,100) {\scriptsize{95.67}};
\node () at (axis cs:20,18,100) {\scriptsize{95.11}};
\node () at (axis cs:4,20,100) {\scriptsize{96.30}};
\node () at (axis cs:6,20,100) {\scriptsize{96.81}};
\node () at (axis cs:8,20,100) {\scriptsize{97.00}};
\node () at (axis cs:10,20,100) {\scriptsize{96.77}};
\node () at (axis cs:12,20,100) {\scriptsize{96.53}};
\node () at (axis cs:14,20,100) {\scriptsize{96.01}};
\node () at (axis cs:16,20,100) {\scriptsize{95.77}};
\node () at (axis cs:18,20,100) {\scriptsize{95.31}};
\node () at (axis cs:20,20,100) {\scriptsize{94.49}};

\draw[very thick,red] (axis cs:17,3,100) -- (axis cs:17,5,100) -- (axis cs:19,5,100) -- (axis cs:19,3,100) -- cycle;
\draw[very thick,red] (axis cs:15,5,100) -- (axis cs:15,7,100) -- (axis cs:17,7,100) -- (axis cs:17,5,100) -- cycle;
\draw[very thick,red] (axis cs:13,7,100) -- (axis cs:13,9,100) -- (axis cs:15,9,100) -- (axis cs:15,7,100) -- cycle;
\draw[very thick,red] (axis cs:9,9,100) -- (axis cs:9,11,100) -- (axis cs:11,11,100) -- (axis cs:11,9,100) -- cycle;
\draw[very thick,red] (axis cs:9,11,100) -- (axis cs:9,13,100) -- (axis cs:11,13,100) -- (axis cs:11,11,100) -- cycle;
\draw[very thick,red] (axis cs:7,13,100) -- (axis cs:7,15,100) -- (axis cs:9,15,100) -- (axis cs:9,13,100) -- cycle;
\draw[very thick,red] (axis cs:7,15,100) -- (axis cs:7,17,100) -- (axis cs:11,17,100) -- (axis cs:11,15,100) -- cycle;
\draw[very thick,red] (axis cs:7,17,100) -- (axis cs:7,19,100) -- (axis cs:9,19,100) -- (axis cs:9,17,100) -- cycle;
\draw[very thick,red] (axis cs:7,19,100) -- (axis cs:7,21,100) -- (axis cs:9,21,100) -- (axis cs:9,19,100) -- cycle;

\draw[very thick,red,densely dotted] (axis cs:3.100000e+00,1.110000e+01,100) -- (axis cs:3.100000e+00,1.290000e+01,100) -- (axis cs:4.900000e+00,1.290000e+01,100) -- (axis cs:4.900000e+00,1.110000e+01,100) -- cycle;
\draw[very thick,red,densely dotted] (axis cs:5.100000e+00,1.910000e+01,100) -- (axis cs:5.100000e+00,2.090000e+01,100) -- (axis cs:6.900000e+00,2.090000e+01,100) -- (axis cs:6.900000e+00,1.910000e+01,100) -- cycle;
\draw[very thick,red,densely dotted] (axis cs:7.100000e+00,1.310000e+01,100) -- (axis cs:7.100000e+00,1.490000e+01,100) -- (axis cs:8.900000e+00,1.490000e+01,100) -- (axis cs:8.900000e+00,1.310000e+01,100) -- cycle;
\draw[very thick,red,densely dotted] (axis cs:9.100000e+00,1.110000e+01,100) -- (axis cs:9.100000e+00,1.290000e+01,100) -- (axis cs:1.090000e+01,1.290000e+01,100) -- (axis cs:1.090000e+01,1.110000e+01,100) -- cycle;
\draw[very thick,red,densely dotted] (axis cs:1.110000e+01,0.710000e+01,100) -- (axis cs:1.110000e+01,0.890000e+01,100) -- (axis cs:1.290000e+01,0.890000e+01,100) -- (axis cs:1.290000e+01,0.710000e+01,100) -- cycle;
\draw[very thick,red,densely dotted] (axis cs:1.310000e+01,7.100000e+00,100) -- (axis cs:1.310000e+01,0.890000e+01,100) -- (axis cs:1.490000e+01,0.890000e+01,100) -- (axis cs:1.490000e+01,7.100000e+00,100) -- cycle;
\draw[very thick,red,densely dotted] (axis cs:1.510000e+01,7.100000e+00,100) -- (axis cs:1.510000e+01,0.890000e+01,100) -- (axis cs:1.690000e+01,0.890000e+01,100) -- (axis cs:1.690000e+01,7.100000e+00,100) -- cycle;
\draw[very thick,red,densely dotted] (axis cs:1.710000e+01,3.100000e+00,100) -- (axis cs:1.710000e+01,4.900000e+00,100) -- (axis cs:1.890000e+01,4.900000e+00,100) -- (axis cs:1.890000e+01,3.100000e+00,100) -- cycle;
\draw[very thick,red,densely dotted] (axis cs:1.910000e+01,5.100000e+00,100) -- (axis cs:1.910000e+01,6.900000e+00,100) -- (axis cs:2.090000e+01,6.900000e+00,100) -- (axis cs:2.090000e+01,5.100000e+00,100) -- cycle;

\draw[very thick,red] (axis cs:3.5,-1.5) rectangle (axis cs:4.5,-0.5);
\node () at (axis cs:12,-1) {\scriptsize{Local optima for certain $s$ and different $T$}};
\draw[very thick,red,densely dotted] (axis cs:3.5,-3) rectangle (axis cs:4.5,-2);
\node () at (axis cs:12,-2.5) {\scriptsize{Local optima for certain $T$ and different $s$}};     
    \end{axis}
\end{tikzpicture}

\caption{Test accuracy with varying \textit{T} and \textit{s}.}
\label{fig:Ts_accuracy}

\end{figure}
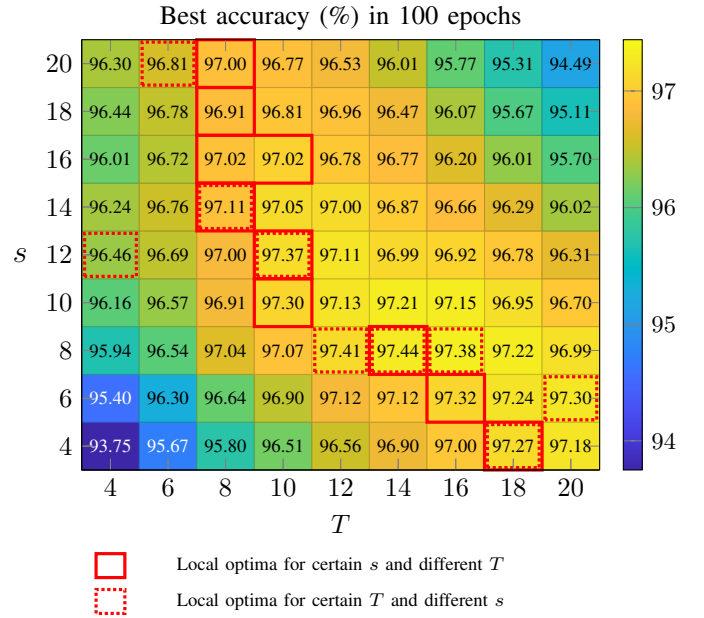

We further explain the observed accuracy trends as follows. As described in Section~\ref{sec:intro}, $s$ controls how strongly the inclusion of a literal is encouraged in Type I feedback. Consequently, increasing (decreasing) $s$ leads the model to capture more (less) fine-grained patterns. This behavior is illustrated in Fig.~\ref{fig:Ts_include}, where the total number of includes generally increases with larger $s$ for a fixed value of $T$. In contrast, $T$ reflects the model's confidence in distinguishing between different classes. Increasing $T$ amplifies the probability of initiating feedback, which in turn encourages the inclusion of more literals. This trend can also be observed in Fig.~\ref{fig:Ts_include}, where the total number of includes increases with larger $T$ for each fixed value of $s$. The granularity of the learned patterns determines the model's generalization capability: excessively fine-grained patterns tend to cause overfitting, whereas overly coarse-grained patterns tend to result in underfitting. Therefore, to mitigate overfitting when $s$ is large, $T$ should be decreased to reduce the probability of feedback. Conversely, when $s$ is small, increasing $T$ is necessary to ensure that the TM learns a sufficient number of included literals, thereby avoiding underfitting.

\pgfplotsset{
	colormap={my colormap}{
rgb=(0.24220000,0.15040000,0.66030000)
rgb=(0.25039048,0.16499524,0.70761429)
rgb=(0.25777143,0.18178095,0.75113810)
rgb=(0.26472857,0.19775714,0.79521429)
rgb=(0.27064762,0.21467619,0.83637143)
rgb=(0.27511429,0.23423810,0.87098571)
rgb=(0.27830000,0.25587143,0.89907143)
rgb=(0.28033333,0.27823333,0.92210000)
rgb=(0.28133810,0.30059524,0.94137619)
rgb=(0.28101429,0.32275714,0.95788571)
rgb=(0.27946667,0.34467143,0.97167619)
rgb=(0.27597143,0.36668095,0.98290476)
rgb=(0.26991429,0.38920000,0.99060000)
rgb=(0.26024286,0.41232857,0.99515714)
rgb=(0.24403333,0.43583333,0.99883333)
rgb=(0.22064286,0.46025714,0.99728571)
rgb=(0.19633333,0.48471905,0.98915238)
rgb=(0.18340476,0.50737143,0.97979524)
rgb=(0.17864286,0.52885714,0.96815714)
rgb=(0.17643810,0.54990476,0.95201905)
rgb=(0.16874286,0.57026190,0.93587143)
rgb=(0.15400000,0.59020000,0.92180000)
rgb=(0.14602857,0.60911905,0.90785714)
rgb=(0.13802381,0.62762857,0.89729048)
rgb=(0.12481429,0.64592857,0.88834286)
rgb=(0.11125238,0.66350000,0.87631429)
rgb=(0.09520952,0.67982857,0.85978095)
rgb=(0.06887143,0.69477143,0.83935714)
rgb=(0.02966667,0.70816667,0.81633333)
rgb=(0.00357143,0.72026667,0.79170000)
rgb=(0.00665714,0.73121429,0.76601429)
rgb=(0.04332857,0.74109524,0.73940952)
rgb=(0.09639524,0.75000000,0.71203810)
rgb=(0.14077143,0.75840000,0.68415714)
rgb=(0.17170000,0.76696190,0.65544286)
rgb=(0.19376667,0.77576667,0.62510000)
rgb=(0.21608571,0.78430000,0.59230000)
rgb=(0.24695714,0.79179524,0.55674286)
rgb=(0.29061429,0.79729048,0.51882857)
rgb=(0.34064286,0.80080000,0.47885714)
rgb=(0.39090000,0.80287143,0.43544762)
rgb=(0.44562857,0.80241905,0.39091905)
rgb=(0.50440000,0.79930000,0.34800000)
rgb=(0.56156190,0.79423333,0.30448095)
rgb=(0.61739524,0.78761905,0.26123810)
rgb=(0.67198571,0.77927143,0.22270000)
rgb=(0.72420000,0.76984286,0.19102857)
rgb=(0.77383333,0.75980476,0.16460952)
rgb=(0.82031429,0.74981429,0.15352857)
rgb=(0.86343333,0.74060000,0.15963333)
rgb=(0.90354286,0.73302857,0.17741429)
rgb=(0.93925714,0.72878571,0.20995714)
rgb=(0.97275714,0.72977143,0.23944286)
rgb=(0.99564762,0.74337143,0.23714762)
rgb=(0.99698571,0.76585714,0.21994286)
rgb=(0.99520476,0.78925238,0.20276190)
rgb=(0.98920000,0.81356667,0.18853333)
rgb=(0.97862857,0.83862857,0.17655714)
rgb=(0.96764762,0.86390000,0.16429048)
rgb=(0.96100952,0.88901905,0.15367619)
rgb=(0.95967143,0.91345714,0.14225714)
rgb=(0.96279524,0.93733810,0.12650952)
rgb=(0.96911429,0.96062857,0.10636190)
rgb=(0.97690000,0.98390000,0.08050000)
	}
}

\begin{figure}[!htb]
	\centering

\begin{tikzpicture}[font=\normalsize, scale=1]
    \begin{axis}[
        xlabel=$T$,
        ylabel=\large{$s$},
        ylabel style={rotate=-90},
        xtick={4,6,8,10,12,14,16,18,20},
        ytick={4,6,8,10,12,14,16,18,20},
        title=Base-10 logarithm of total includes,
        title style={yshift=-0.2cm},
        view={0}{90},
        colorbar,colormap access=map,
        colormap name = my colormap,
        colorbar/width=2.5mm,
        colorbar style={font=\small,xshift=0cm},
        clip=false
    ]
        \addplot3 [surf,mesh/cols=10] table {Shengyu/Data/Ts_include.dat};

\node () at (axis cs:4,4,100) {\scriptsize{\textcolor{white}{5.09}}};
\node () at (axis cs:6,4,100) {\scriptsize{\textcolor{white}{5.00}}};
\node () at (axis cs:8,4,100) {\scriptsize{\textcolor{white}{4.94}}};
\node () at (axis cs:10,4,100) {\scriptsize{\textcolor{white}{4.92}}};
\node () at (axis cs:12,4,100) {\scriptsize{\textcolor{white}{4.90}}};
\node () at (axis cs:14,4,100) {\scriptsize{\textcolor{white}{4.93}}};
\node () at (axis cs:16,4,100) {\scriptsize{\textcolor{white}{5.06}}};
\node () at (axis cs:18,4,100) {\scriptsize{\textcolor{white}{5.15}}};
\node () at (axis cs:20,4,100) {\scriptsize{\textcolor{white}{5.20}}};
\node () at (axis cs:4,6,100) {\scriptsize{\textcolor{white}{4.99}}};
\node () at (axis cs:6,6,100) {\scriptsize{\textcolor{white}{4.91}}};
\node () at (axis cs:8,6,100) {\scriptsize{\textcolor{white}{4.82}}};
\node () at (axis cs:10,6,100) {\scriptsize{\textcolor{white}{4.85}}};
\node () at (axis cs:12,6,100) {\scriptsize{\textcolor{white}{4.91}}};
\node () at (axis cs:14,6,100) {\scriptsize{\textcolor{white}{4.97}}};
\node () at (axis cs:16,6,100) {\scriptsize{\textcolor{white}{5.06}}};
\node () at (axis cs:18,6,100) {\scriptsize{\textcolor{white}{5.17}}};
\node () at (axis cs:20,6,100) {\scriptsize{\textcolor{white}{5.27}}};
\node () at (axis cs:4,8,100) {\scriptsize{\textcolor{white}{4.90}}};
\node () at (axis cs:6,8,100) {\scriptsize{\textcolor{white}{4.87}}};
\node () at (axis cs:8,8,100) {\scriptsize{\textcolor{white}{4.86}}};
\node () at (axis cs:10,8,100) {\scriptsize{\textcolor{white}{4.92}}};
\node () at (axis cs:12,8,100) {\scriptsize{\textcolor{white}{4.98}}};
\node () at (axis cs:14,8,100) {\scriptsize{\textcolor{white}{5.09}}};
\node () at (axis cs:16,8,100) {\scriptsize{\textcolor{white}{5.20}}};
\node () at (axis cs:18,8,100) {\scriptsize{\textcolor{white}{5.31}}};
\node () at (axis cs:20,8,100) {\scriptsize{5.42}};
\node () at (axis cs:4,10,100) {\scriptsize{\textcolor{white}{4.91}}};
\node () at (axis cs:6,10,100) {\scriptsize{\textcolor{white}{4.84}}};
\node () at (axis cs:8,10,100) {\scriptsize{\textcolor{white}{4.88}}};
\node () at (axis cs:10,10,100) {\scriptsize{\textcolor{white}{4.97}}};
\node () at (axis cs:12,10,100) {\scriptsize{\textcolor{white}{5.04}}};
\node () at (axis cs:14,10,100) {\scriptsize{\textcolor{white}{5.16}}};
\node () at (axis cs:16,10,100) {\scriptsize{\textcolor{white}{5.34}}};
\node () at (axis cs:18,10,100) {\scriptsize{5.39}};
\node () at (axis cs:20,10,100) {\scriptsize{5.57}};
\node () at (axis cs:4,12,100) {\scriptsize{\textcolor{white}{4.89}}};
\node () at (axis cs:6,12,100) {\scriptsize{\textcolor{white}{4.89}}};
\node () at (axis cs:8,12,100) {\scriptsize{\textcolor{white}{4.94}}};
\node () at (axis cs:10,12,100) {\scriptsize{\textcolor{white}{5.06}}};
\node () at (axis cs:12,12,100) {\scriptsize{\textcolor{white}{5.18}}};
\node () at (axis cs:14,12,100) {\scriptsize{\textcolor{white}{5.35}}};
\node () at (axis cs:16,12,100) {\scriptsize{5.49}};
\node () at (axis cs:18,12,100) {\scriptsize{5.56}};
\node () at (axis cs:20,12,100) {\scriptsize{5.66}};
\node () at (axis cs:4,14,100) {\scriptsize{\textcolor{white}{4.89}}};
\node () at (axis cs:6,14,100) {\scriptsize{\textcolor{white}{4.91}}};
\node () at (axis cs:8,14,100) {\scriptsize{\textcolor{white}{5.00}}};
\node () at (axis cs:10,14,100) {\scriptsize{\textcolor{white}{5.16}}};
\node () at (axis cs:12,14,100) {\scriptsize{\textcolor{white}{5.31}}};
\node () at (axis cs:14,14,100) {\scriptsize{5.42}};
\node () at (axis cs:16,14,100) {\scriptsize{5.57}};
\node () at (axis cs:18,14,100) {\scriptsize{5.61}};
\node () at (axis cs:20,14,100) {\scriptsize{5.76}};
\node () at (axis cs:4,16,100) {\scriptsize{\textcolor{white}{4.90}}};
\node () at (axis cs:6,16,100) {\scriptsize{\textcolor{white}{4.97}}};
\node () at (axis cs:8,16,100) {\scriptsize{\textcolor{white}{5.05}}};
\node () at (axis cs:10,16,100) {\scriptsize{\textcolor{white}{5.24}}};
\node () at (axis cs:12,16,100) {\scriptsize{5.40}};
\node () at (axis cs:14,16,100) {\scriptsize{5.42}};
\node () at (axis cs:16,16,100) {\scriptsize{5.67}};
\node () at (axis cs:18,16,100) {\scriptsize{5.73}};
\node () at (axis cs:20,16,100) {\scriptsize{5.82}};
\node () at (axis cs:4,18,100) {\scriptsize{\textcolor{white}{4.89}}};
\node () at (axis cs:6,18,100) {\scriptsize{\textcolor{white}{5.02}}};
\node () at (axis cs:8,18,100) {\scriptsize{\textcolor{white}{5.15}}};
\node () at (axis cs:10,18,100) {\scriptsize{\textcolor{white}{5.27}}};
\node () at (axis cs:12,18,100) {\scriptsize{5.47}};
\node () at (axis cs:14,18,100) {\scriptsize{5.64}};
\node () at (axis cs:16,18,100) {\scriptsize{5.72}};
\node () at (axis cs:18,18,100) {\scriptsize{5.77}};
\node () at (axis cs:20,18,100) {\scriptsize{5.81}};
\node () at (axis cs:4,20,100) {\scriptsize{\textcolor{white}{4.95}}};
\node () at (axis cs:6,20,100) {\scriptsize{\textcolor{white}{5.03}}};
\node () at (axis cs:8,20,100) {\scriptsize{\textcolor{white}{5.21}}};
\node () at (axis cs:10,20,100) {\scriptsize{5.39}};
\node () at (axis cs:12,20,100) {\scriptsize{5.49}};
\node () at (axis cs:14,20,100) {\scriptsize{5.64}};
\node () at (axis cs:16,20,100) {\scriptsize{5.78}};
\node () at (axis cs:18,20,100) {\scriptsize{5.87}};
\node () at (axis cs:20,20,100) {\scriptsize{5.88}};
          
    \end{axis}
\end{tikzpicture}

\caption{Number of includes with varying \textit{T} and \textit{s}. Too many includes lead to overfitting, while too few lead to underfitting.}
\label{fig:Ts_include}

\end{figure}
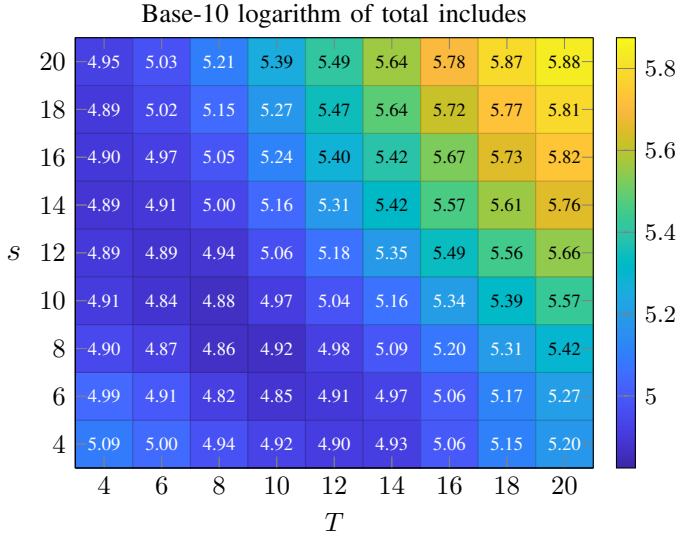

\subsection{Impact of $T$ and $s$ on Inference Latency}  \label{subsec:ts_latency}
The TM models corresponding to all ($T$,$s$) combinations are implemented using the software framework described in Section~\ref{sec:software}. We evaluate the inference latency of each model by measuring the average number of clock cycles over 100 data points on an ARM architecture simulated with gem5 \cite{10.1145/2024716.2024718}. The results are presented in Fig.~\ref{fig:Ts_latency}. As shown, the inference latency decreases noticeably and consistently with increasing $s$ for each fixed value of $T$. In contrast, the latency is relatively insensitive to changes in $T$ for a fixed value of $s$, as evidenced by the small variation across each row.

\pgfplotsset{
	colormap={my colormap}{
rgb=(0.24220000,0.15040000,0.66030000)
rgb=(0.25039048,0.16499524,0.70761429)
rgb=(0.25777143,0.18178095,0.75113810)
rgb=(0.26472857,0.19775714,0.79521429)
rgb=(0.27064762,0.21467619,0.83637143)
rgb=(0.27511429,0.23423810,0.87098571)
rgb=(0.27830000,0.25587143,0.89907143)
rgb=(0.28033333,0.27823333,0.92210000)
rgb=(0.28133810,0.30059524,0.94137619)
rgb=(0.28101429,0.32275714,0.95788571)
rgb=(0.27946667,0.34467143,0.97167619)
rgb=(0.27597143,0.36668095,0.98290476)
rgb=(0.26991429,0.38920000,0.99060000)
rgb=(0.26024286,0.41232857,0.99515714)
rgb=(0.24403333,0.43583333,0.99883333)
rgb=(0.22064286,0.46025714,0.99728571)
rgb=(0.19633333,0.48471905,0.98915238)
rgb=(0.18340476,0.50737143,0.97979524)
rgb=(0.17864286,0.52885714,0.96815714)
rgb=(0.17643810,0.54990476,0.95201905)
rgb=(0.16874286,0.57026190,0.93587143)
rgb=(0.15400000,0.59020000,0.92180000)
rgb=(0.14602857,0.60911905,0.90785714)
rgb=(0.13802381,0.62762857,0.89729048)
rgb=(0.12481429,0.64592857,0.88834286)
rgb=(0.11125238,0.66350000,0.87631429)
rgb=(0.09520952,0.67982857,0.85978095)
rgb=(0.06887143,0.69477143,0.83935714)
rgb=(0.02966667,0.70816667,0.81633333)
rgb=(0.00357143,0.72026667,0.79170000)
rgb=(0.00665714,0.73121429,0.76601429)
rgb=(0.04332857,0.74109524,0.73940952)
rgb=(0.09639524,0.75000000,0.71203810)
rgb=(0.14077143,0.75840000,0.68415714)
rgb=(0.17170000,0.76696190,0.65544286)
rgb=(0.19376667,0.77576667,0.62510000)
rgb=(0.21608571,0.78430000,0.59230000)
rgb=(0.24695714,0.79179524,0.55674286)
rgb=(0.29061429,0.79729048,0.51882857)
rgb=(0.34064286,0.80080000,0.47885714)
rgb=(0.39090000,0.80287143,0.43544762)
rgb=(0.44562857,0.80241905,0.39091905)
rgb=(0.50440000,0.79930000,0.34800000)
rgb=(0.56156190,0.79423333,0.30448095)
rgb=(0.61739524,0.78761905,0.26123810)
rgb=(0.67198571,0.77927143,0.22270000)
rgb=(0.72420000,0.76984286,0.19102857)
rgb=(0.77383333,0.75980476,0.16460952)
rgb=(0.82031429,0.74981429,0.15352857)
rgb=(0.86343333,0.74060000,0.15963333)
rgb=(0.90354286,0.73302857,0.17741429)
rgb=(0.93925714,0.72878571,0.20995714)
rgb=(0.97275714,0.72977143,0.23944286)
rgb=(0.99564762,0.74337143,0.23714762)
rgb=(0.99698571,0.76585714,0.21994286)
rgb=(0.99520476,0.78925238,0.20276190)
rgb=(0.98920000,0.81356667,0.18853333)
rgb=(0.97862857,0.83862857,0.17655714)
rgb=(0.96764762,0.86390000,0.16429048)
rgb=(0.96100952,0.88901905,0.15367619)
rgb=(0.95967143,0.91345714,0.14225714)
rgb=(0.96279524,0.93733810,0.12650952)
rgb=(0.96911429,0.96062857,0.10636190)
rgb=(0.97690000,0.98390000,0.08050000)
	}
}

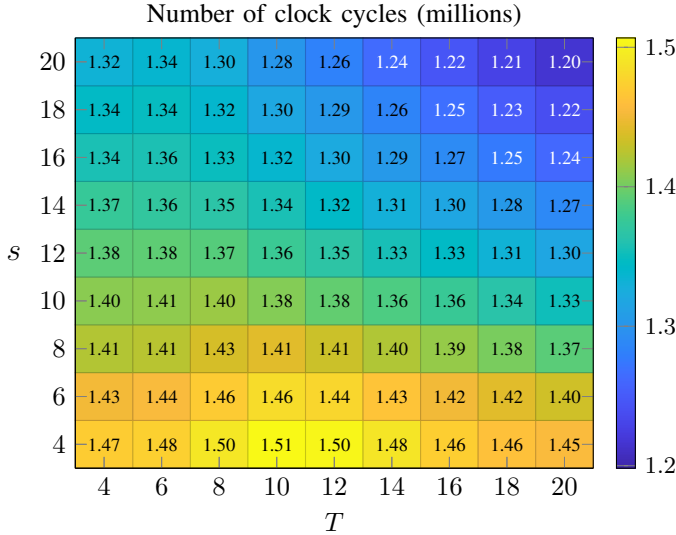
\begin{figure}[!htb]
	\centering

\begin{tikzpicture}[font=\normalsize, scale=1]
    \begin{axis}[
        xlabel=$T$,
        ylabel=\large{$s$},
        ylabel style={rotate=-90},
        xtick={4,6,8,10,12,14,16,18,20},
        ytick={4,6,8,10,12,14,16,18,20},
        title=Number of clock cycles (millions),
        title style={yshift=-0.2cm},
        view={0}{90},
        colorbar,colormap access=map,
        colormap name = my colormap,
        colorbar/width=2.5mm,
        colorbar style={font=\small,xshift=0cm},
        clip=false
    ]
        \addplot3 [surf,mesh/cols=10] table {Shengyu/Data/Ts_latency.dat};
 
\node () at (axis cs:4,4,1.5) {\scriptsize{1.47}};
\node () at (axis cs:6,4,1.5) {\scriptsize{1.48}};
\node () at (axis cs:8,4,1.5) {\scriptsize{1.50}};
\node () at (axis cs:10,4,1.5) {\scriptsize{1.51}};
\node () at (axis cs:12,4,1.5) {\scriptsize{1.50}};
\node () at (axis cs:14,4,1.5) {\scriptsize{1.48}};
\node () at (axis cs:16,4,1.5) {\scriptsize{1.46}};
\node () at (axis cs:18,4,1.5) {\scriptsize{1.46}};
\node () at (axis cs:20,4,1.5) {\scriptsize{1.45}};
\node () at (axis cs:4,6,1.5) {\scriptsize{1.43}};
\node () at (axis cs:6,6,1.5) {\scriptsize{1.44}};
\node () at (axis cs:8,6,1.5) {\scriptsize{1.46}};
\node () at (axis cs:10,6,1.5) {\scriptsize{1.46}};
\node () at (axis cs:12,6,1.5) {\scriptsize{1.44}};
\node () at (axis cs:14,6,1.5) {\scriptsize{1.43}};
\node () at (axis cs:16,6,1.5) {\scriptsize{1.42}};
\node () at (axis cs:18,6,1.5) {\scriptsize{1.42}};
\node () at (axis cs:20,6,1.5) {\scriptsize{1.40}};
\node () at (axis cs:4,8,1.5) {\scriptsize{1.41}};
\node () at (axis cs:6,8,1.5) {\scriptsize{1.41}};
\node () at (axis cs:8,8,1.5) {\scriptsize{1.43}};
\node () at (axis cs:10,8,1.5) {\scriptsize{1.41}};
\node () at (axis cs:12,8,1.5) {\scriptsize{1.41}};
\node () at (axis cs:14,8,1.5) {\scriptsize{1.40}};
\node () at (axis cs:16,8,1.5) {\scriptsize{1.39}};
\node () at (axis cs:18,8,1.5) {\scriptsize{1.38}};
\node () at (axis cs:20,8,1.5) {\scriptsize{1.37}};
\node () at (axis cs:4,10,1.5) {\scriptsize{1.40}};
\node () at (axis cs:6,10,1.5) {\scriptsize{1.41}};
\node () at (axis cs:8,10,1.5) {\scriptsize{1.40}};
\node () at (axis cs:10,10,1.5) {\scriptsize{1.38}};
\node () at (axis cs:12,10,1.5) {\scriptsize{1.38}};
\node () at (axis cs:14,10,1.5) {\scriptsize{1.36}};
\node () at (axis cs:16,10,1.5) {\scriptsize{1.36}};
\node () at (axis cs:18,10,1.5) {\scriptsize{1.34}};
\node () at (axis cs:20,10,1.5) {\scriptsize{1.33}};
\node () at (axis cs:4,12,1.5) {\scriptsize{1.38}};
\node () at (axis cs:6,12,1.5) {\scriptsize{1.38}};
\node () at (axis cs:8,12,1.5) {\scriptsize{1.37}};
\node () at (axis cs:10,12,1.5) {\scriptsize{1.36}};
\node () at (axis cs:12,12,1.5) {\scriptsize{1.35}};
\node () at (axis cs:14,12,1.5) {\scriptsize{1.33}};
\node () at (axis cs:16,12,1.5) {\scriptsize{1.33}};
\node () at (axis cs:18,12,1.5) {\scriptsize{1.31}};
\node () at (axis cs:20,12,1.5) {\scriptsize{1.30}};
\node () at (axis cs:4,14,1.5) {\scriptsize{1.37}};
\node () at (axis cs:6,14,1.5) {\scriptsize{1.36}};
\node () at (axis cs:8,14,1.5) {\scriptsize{1.35}};
\node () at (axis cs:10,14,1.5) {\scriptsize{1.34}};
\node () at (axis cs:12,14,1.5) {\scriptsize{1.32}};
\node () at (axis cs:14,14,1.5) {\scriptsize{1.31}};
\node () at (axis cs:16,14,1.5) {\scriptsize{1.30}};
\node () at (axis cs:18,14,1.5) {\scriptsize{1.28}};
\node () at (axis cs:20,14,1.5) {\scriptsize{1.27}};
\node () at (axis cs:4,16,1.5) {\scriptsize{1.34}};
\node () at (axis cs:6,16,1.5) {\scriptsize{1.36}};
\node () at (axis cs:8,16,1.5) {\scriptsize{1.33}};
\node () at (axis cs:10,16,1.5) {\scriptsize{1.32}};
\node () at (axis cs:12,16,1.5) {\scriptsize{1.30}};
\node () at (axis cs:14,16,1.5) {\scriptsize{1.29}};
\node () at (axis cs:16,16,1.5) {\scriptsize{1.27}};
\node () at (axis cs:18,16,1.5) {\scriptsize{\textcolor{white}{1.25}}};
\node () at (axis cs:20,16,1.5) {\scriptsize{\textcolor{white}{1.24}}};
\node () at (axis cs:4,18,1.5) {\scriptsize{1.34}};
\node () at (axis cs:6,18,1.5) {\scriptsize{1.34}};
\node () at (axis cs:8,18,1.5) {\scriptsize{1.32}};
\node () at (axis cs:10,18,1.5) {\scriptsize{1.30}};
\node () at (axis cs:12,18,1.5) {\scriptsize{1.29}};
\node () at (axis cs:14,18,1.5) {\scriptsize{1.26}};
\node () at (axis cs:16,18,1.5) {\scriptsize{\textcolor{white}{1.25}}};
\node () at (axis cs:18,18,1.5) {\scriptsize{\textcolor{white}{1.23}}};
\node () at (axis cs:20,18,1.5) {\scriptsize{\textcolor{white}{1.22}}};
\node () at (axis cs:4,20,1.5) {\scriptsize{1.32}};
\node () at (axis cs:6,20,1.5) {\scriptsize{1.34}};
\node () at (axis cs:8,20,1.5) {\scriptsize{1.30}};
\node () at (axis cs:10,20,1.5) {\scriptsize{1.28}};
\node () at (axis cs:12,20,1.5) {\scriptsize{1.26}};
\node () at (axis cs:14,20,1.5) {\scriptsize{\textcolor{white}{1.24}}};
\node () at (axis cs:16,20,1.5) {\scriptsize{\textcolor{white}{1.22}}};
\node () at (axis cs:18,20,1.5) {\scriptsize{\textcolor{white}{1.21}}};
\node () at (axis cs:20,20,1.5) {\scriptsize{\textcolor{white}{1.20}}};
 
      \end{axis}
     
\end{tikzpicture}

\caption{Inference latency with varying \textit{T} and \textit{s}. Latency tends to decrease as \textit{s} increases.}
\label{fig:Ts_latency}

\end{figure}

For the implemented software, inference latency for a given datapoint is determined by two factors: (1) the total number of votes, i.e., the number of clauses producing an output of 1, and (2) how early the evaluation of clauses producing an output of 0 can be terminated. Specifically, a larger number of votes increases inference latency because more clauses must be evaluated to completion for pattern recognition, requiring comparisons across all features without the possibility of early termination. In contrast, the earlier a mismatch is detected within a clause yielding 0, the sooner its evaluation can terminate, reducing the overall inference latency.

We first measure the average number of votes over 100 datapoints for different values of $T$ and $s$, as shown in Fig.~\ref{fig:Ts_vote}. It is evident that the number of votes decreases along each column, indicating that larger values of $s$ cause fewer clauses to produce a vote. This is because a larger $s$ results in more fine-grained clause representations, with each clause containing more literals. Consequently, the learned patterns become more specific and are therefore less likely to match unseen input data, leading to fewer clauses producing an output of 1.
Meanwhile, along each row, larger values of $T$ lead to a greater number of votes. This behavior stems from the TM feedback mechanism, which reinforces clause expressions (learned patterns) until the class sum approaches $T$ or -$T$. As a result, increasing $T$ allows more clauses to be reinforced, potentially leading to a larger number of voting clauses. However, the empirical results in Fig.~\ref{fig:Ts_vote} indicate that the number of votes is considerably more sensitive to changes in $s$ than to changes in $T$.

\pgfplotsset{
	colormap={my colormap}{
rgb=(0.24220000,0.15040000,0.66030000)
rgb=(0.25039048,0.16499524,0.70761429)
rgb=(0.25777143,0.18178095,0.75113810)
rgb=(0.26472857,0.19775714,0.79521429)
rgb=(0.27064762,0.21467619,0.83637143)
rgb=(0.27511429,0.23423810,0.87098571)
rgb=(0.27830000,0.25587143,0.89907143)
rgb=(0.28033333,0.27823333,0.92210000)
rgb=(0.28133810,0.30059524,0.94137619)
rgb=(0.28101429,0.32275714,0.95788571)
rgb=(0.27946667,0.34467143,0.97167619)
rgb=(0.27597143,0.36668095,0.98290476)
rgb=(0.26991429,0.38920000,0.99060000)
rgb=(0.26024286,0.41232857,0.99515714)
rgb=(0.24403333,0.43583333,0.99883333)
rgb=(0.22064286,0.46025714,0.99728571)
rgb=(0.19633333,0.48471905,0.98915238)
rgb=(0.18340476,0.50737143,0.97979524)
rgb=(0.17864286,0.52885714,0.96815714)
rgb=(0.17643810,0.54990476,0.95201905)
rgb=(0.16874286,0.57026190,0.93587143)
rgb=(0.15400000,0.59020000,0.92180000)
rgb=(0.14602857,0.60911905,0.90785714)
rgb=(0.13802381,0.62762857,0.89729048)
rgb=(0.12481429,0.64592857,0.88834286)
rgb=(0.11125238,0.66350000,0.87631429)
rgb=(0.09520952,0.67982857,0.85978095)
rgb=(0.06887143,0.69477143,0.83935714)
rgb=(0.02966667,0.70816667,0.81633333)
rgb=(0.00357143,0.72026667,0.79170000)
rgb=(0.00665714,0.73121429,0.76601429)
rgb=(0.04332857,0.74109524,0.73940952)
rgb=(0.09639524,0.75000000,0.71203810)
rgb=(0.14077143,0.75840000,0.68415714)
rgb=(0.17170000,0.76696190,0.65544286)
rgb=(0.19376667,0.77576667,0.62510000)
rgb=(0.21608571,0.78430000,0.59230000)
rgb=(0.24695714,0.79179524,0.55674286)
rgb=(0.29061429,0.79729048,0.51882857)
rgb=(0.34064286,0.80080000,0.47885714)
rgb=(0.39090000,0.80287143,0.43544762)
rgb=(0.44562857,0.80241905,0.39091905)
rgb=(0.50440000,0.79930000,0.34800000)
rgb=(0.56156190,0.79423333,0.30448095)
rgb=(0.61739524,0.78761905,0.26123810)
rgb=(0.67198571,0.77927143,0.22270000)
rgb=(0.72420000,0.76984286,0.19102857)
rgb=(0.77383333,0.75980476,0.16460952)
rgb=(0.82031429,0.74981429,0.15352857)
rgb=(0.86343333,0.74060000,0.15963333)
rgb=(0.90354286,0.73302857,0.17741429)
rgb=(0.93925714,0.72878571,0.20995714)
rgb=(0.97275714,0.72977143,0.23944286)
rgb=(0.99564762,0.74337143,0.23714762)
rgb=(0.99698571,0.76585714,0.21994286)
rgb=(0.99520476,0.78925238,0.20276190)
rgb=(0.98920000,0.81356667,0.18853333)
rgb=(0.97862857,0.83862857,0.17655714)
rgb=(0.96764762,0.86390000,0.16429048)
rgb=(0.96100952,0.88901905,0.15367619)
rgb=(0.95967143,0.91345714,0.14225714)
rgb=(0.96279524,0.93733810,0.12650952)
rgb=(0.96911429,0.96062857,0.10636190)
rgb=(0.97690000,0.98390000,0.08050000)
	}
}

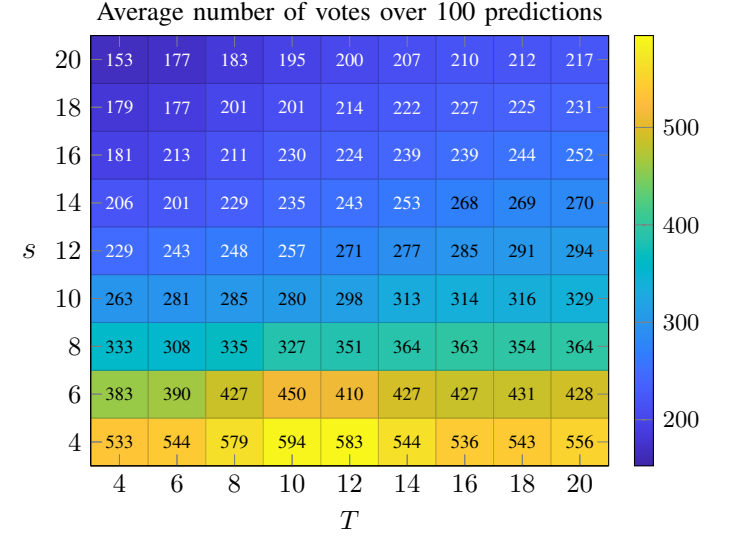
\begin{figure}[!htb]
	\centering

\begin{tikzpicture}[font=\normalsize, scale=1]
    \begin{axis}[
        xlabel=$T$,
        ylabel=\large{$s$},
        ylabel style={rotate=-90},
        xtick={4,6,8,10,12,14,16,18,20},
        ytick={4,6,8,10,12,14,16,18,20},
        title=Average number of votes over 100 predictions,
        title style={yshift=-0.2cm},
        view={0}{90},
        colorbar,colormap access=map,
        colormap name = my colormap,
        colorbar/width=2.5mm,
        colorbar style={font=\small,xshift=0cm},
        clip=false
    ]
        \addplot3 [surf,mesh/cols=10] table {Shengyu/Data/Ts_vote.dat};

\node () at (axis cs:4,4,100) {\scriptsize{533}};
\node () at (axis cs:6,4,100) {\scriptsize{544}};
\node () at (axis cs:8,4,100) {\scriptsize{579}};
\node () at (axis cs:10,4,100) {\scriptsize{594}};
\node () at (axis cs:12,4,100) {\scriptsize{583}};
\node () at (axis cs:14,4,100) {\scriptsize{544}};
\node () at (axis cs:16,4,100) {\scriptsize{536}};
\node () at (axis cs:18,4,100) {\scriptsize{543}};
\node () at (axis cs:20,4,100) {\scriptsize{556}};
\node () at (axis cs:4,6,100) {\scriptsize{383}};
\node () at (axis cs:6,6,100) {\scriptsize{390}};
\node () at (axis cs:8,6,100) {\scriptsize{427}};
\node () at (axis cs:10,6,100) {\scriptsize{450}};
\node () at (axis cs:12,6,100) {\scriptsize{410}};
\node () at (axis cs:14,6,100) {\scriptsize{427}};
\node () at (axis cs:16,6,100) {\scriptsize{427}};
\node () at (axis cs:18,6,100) {\scriptsize{431}};
\node () at (axis cs:20,6,100) {\scriptsize{428}};
\node () at (axis cs:4,8,100) {\scriptsize{333}};
\node () at (axis cs:6,8,100) {\scriptsize{308}};
\node () at (axis cs:8,8,100) {\scriptsize{335}};
\node () at (axis cs:10,8,100) {\scriptsize{327}};
\node () at (axis cs:12,8,100) {\scriptsize{351}};
\node () at (axis cs:14,8,100) {\scriptsize{364}};
\node () at (axis cs:16,8,100) {\scriptsize{363}};
\node () at (axis cs:18,8,100) {\scriptsize{354}};
\node () at (axis cs:20,8,100) {\scriptsize{364}};
\node () at (axis cs:4,10,100) {\scriptsize{263}};
\node () at (axis cs:6,10,100) {\scriptsize{281}};
\node () at (axis cs:8,10,100) {\scriptsize{285}};
\node () at (axis cs:10,10,100) {\scriptsize{280}};
\node () at (axis cs:12,10,100) {\scriptsize{298}};
\node () at (axis cs:14,10,100) {\scriptsize{313}};
\node () at (axis cs:16,10,100) {\scriptsize{314}};
\node () at (axis cs:18,10,100) {\scriptsize{316}};
\node () at (axis cs:20,10,100) {\scriptsize{329}};
\node () at (axis cs:4,12,100) {\scriptsize{\textcolor{white}{229}}};
\node () at (axis cs:6,12,100) {\scriptsize{\textcolor{white}{243}}};
\node () at (axis cs:8,12,100) {\scriptsize{\textcolor{white}{248}}};
\node () at (axis cs:10,12,100) {\scriptsize{\textcolor{white}{257}}};
\node () at (axis cs:12,12,100) {\scriptsize{271}};
\node () at (axis cs:14,12,100) {\scriptsize{277}};
\node () at (axis cs:16,12,100) {\scriptsize{285}};
\node () at (axis cs:18,12,100) {\scriptsize{291}};
\node () at (axis cs:20,12,100) {\scriptsize{294}};
\node () at (axis cs:4,14,100) {\scriptsize{\textcolor{white}{206}}};
\node () at (axis cs:6,14,100) {\scriptsize{\textcolor{white}{201}}};
\node () at (axis cs:8,14,100) {\scriptsize{\textcolor{white}{229}}};
\node () at (axis cs:10,14,100) {\scriptsize{\textcolor{white}{235}}};
\node () at (axis cs:12,14,100) {\scriptsize{\textcolor{white}{243}}};
\node () at (axis cs:14,14,100) {\scriptsize{\textcolor{white}{253}}};
\node () at (axis cs:16,14,100) {\scriptsize{268}};
\node () at (axis cs:18,14,100) {\scriptsize{269}};
\node () at (axis cs:20,14,100) {\scriptsize{270}};
\node () at (axis cs:4,16,100) {\scriptsize{\textcolor{white}{181}}};
\node () at (axis cs:6,16,100) {\scriptsize{\textcolor{white}{213}}};
\node () at (axis cs:8,16,100) {\scriptsize{\textcolor{white}{211}}};
\node () at (axis cs:10,16,100) {\scriptsize{\textcolor{white}{230}}};
\node () at (axis cs:12,16,100) {\scriptsize{\textcolor{white}{224}}};
\node () at (axis cs:14,16,100) {\scriptsize{\textcolor{white}{239}}};
\node () at (axis cs:16,16,100) {\scriptsize{\textcolor{white}{239}}};
\node () at (axis cs:18,16,100) {\scriptsize{\textcolor{white}{244}}};
\node () at (axis cs:20,16,100) {\scriptsize{\textcolor{white}{252}}};
\node () at (axis cs:4,18,100) {\scriptsize{\textcolor{white}{179}}};
\node () at (axis cs:6,18,100) {\scriptsize{\textcolor{white}{177}}};
\node () at (axis cs:8,18,100) {\scriptsize{\textcolor{white}{201}}};
\node () at (axis cs:10,18,100) {\scriptsize{\textcolor{white}{201}}};
\node () at (axis cs:12,18,100) {\scriptsize{\textcolor{white}{214}}};
\node () at (axis cs:14,18,100) {\scriptsize{\textcolor{white}{222}}};
\node () at (axis cs:16,18,100) {\scriptsize{\textcolor{white}{227}}};
\node () at (axis cs:18,18,100) {\scriptsize{\textcolor{white}{225}}};
\node () at (axis cs:20,18,100) {\scriptsize{\textcolor{white}{231}}};
\node () at (axis cs:4,20,100) {\scriptsize{\textcolor{white}{153}}};
\node () at (axis cs:6,20,100) {\scriptsize{\textcolor{white}{177}}};
\node () at (axis cs:8,20,100) {\scriptsize{\textcolor{white}{183}}};
\node () at (axis cs:10,20,100) {\scriptsize{\textcolor{white}{195}}};
\node () at (axis cs:12,20,100) {\scriptsize{\textcolor{white}{200}}};
\node () at (axis cs:14,20,100) {\scriptsize{\textcolor{white}{207}}};
\node () at (axis cs:16,20,100) {\scriptsize{\textcolor{white}{210}}};
\node () at (axis cs:18,20,100) {\scriptsize{\textcolor{white}{212}}};
\node () at (axis cs:20,20,100) {\scriptsize{\textcolor{white}{217}}};
 
      \end{axis}
     
\end{tikzpicture}

\caption{Number of votes (clauses producing output 1) with varying \textit{T} and \textit{s}. Vote count tends to decrease as \textit{s} increases.}
\label{fig:Ts_vote}

\end{figure}

We next examine the second factor, i.e., how early the evaluation of a
clause producing an output of 0 can be terminated. Recall from
Section~\ref{sec:software} that early exit is triggered at the first feature that is
included and mismatched. For a given dataset, the
probability that a feature takes the value 0 at a given position is an
intrinsic property of the data and is therefore a constant, independent
of $T$ and $s$. Consequently, the probability that any particular
position triggers an exit is governed solely by the probability that
the corresponding feature is included. As shown in Fig.~\ref{fig:Ts_include}, increasing
either $T$ or $s$ raises the number of includes and hence the
per-feature include probability. A higher include probability makes
each evaluated position more likely to carry a mismatch, so the exit is expected to be triggered earlier in the
evaluation sequence.

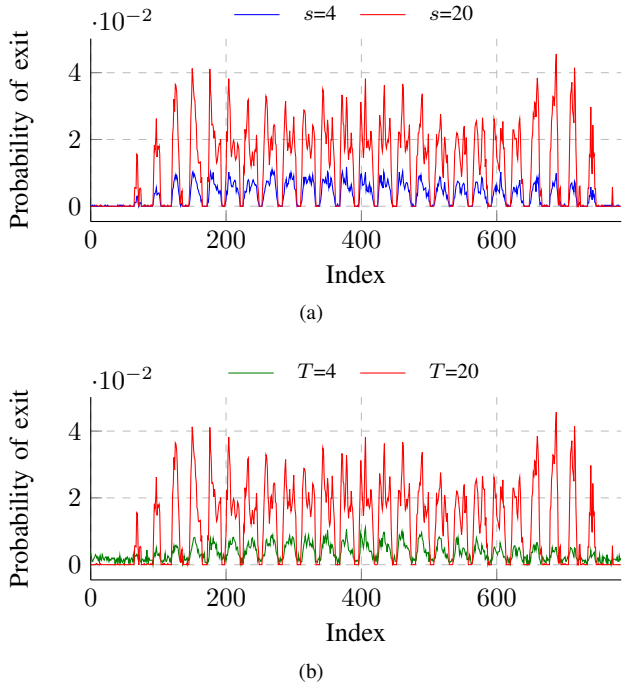
\begin{figure}[t]
\centering

\subfloat[]{
\begin{tikzpicture}[font=\normalsize, scale=1]
\begin{axis}[
height=4cm,
width=8.6cm,
axis x line*=bottom,
axis y line*=left,
xmin=0, xmax=784,
grid=major,
grid style={dashed,gray!50},
xlabel = Index,
ylabel = Probability of exit,
ylabel style={yshift=-2ex},
xlabel style={yshift=0.5ex},
clip=false,
legend style={at={(0.5,1.02)},anchor=south,legend columns=2,
              font=\footnotesize,draw=none,column sep=1.5ex}
]
\addplot [blue] table [y=t20s4, x=idx] {Yefan/Data/prob_ts.dat};
\addlegendentry{$s$=4}
\addplot [red] table [y=t20s20, x=idx] {Yefan/Data/prob_ts.dat};
\addlegendentry{$s$=20}
\end{axis}
\end{tikzpicture}
\label{fig:prob_ts_a}
}
\\
\subfloat[]{
\begin{tikzpicture}[font=\normalsize, scale=1]
\begin{axis}[
height=4cm,
width=8.6cm,
axis x line*=bottom,
axis y line*=left,
xmin=0, xmax=784,
grid=major,
grid style={dashed,gray!50},
xlabel = Index,
ylabel = Probability of exit,
ylabel style={yshift=-2ex},
xlabel style={yshift=0.5ex},
clip=false,
legend style={at={(0.5,1.02)},anchor=south,legend columns=2,
              font=\footnotesize,draw=none,column sep=1.5ex}
]
\addplot [green!50!black] table [y=t4s20, x=idx] {Yefan/Data/prob_ts.dat};
\addlegendentry{$T$=4}
\addplot [red] table [y=t20s20, x=idx] {Yefan/Data/prob_ts.dat};
\addlegendentry{$T$=20}
\end{axis}
\end{tikzpicture}
\label{fig:prob_ts_b}
}

	\caption{Probability of terminating clause computation at different execution indices for (a) fixed $T$ = 20 with varying $s$, and (b) fixed $s$ = 20 with varying $T$. The execution index denotes the order of computation rather than the feature index. Data points are connected for visual clarity.
 }
	\label{fig:prob_ts}
\end{figure}

To verify this, we measure the probability of terminating clause
computation at each execution index for different $(T,s)$ combinations,
as reported in Fig.~\ref{fig:prob_ts}. Here the execution index denotes the order in
which literals are evaluated rather than the feature index, so a
distribution concentrated at small indices indicates frequent and early
termination. In Fig.~\ref{fig:prob_ts_a}, $T$ is fixed and $s$ is increased: the exit
probability at the early indices rises markedly, confirming that a
larger $s$ leads to earlier termination. In Fig.~\ref{fig:prob_ts_b}, $s$ is fixed and
$T$ is increased, which likewise raises the exit probability at the
early indices. Both hyperparameters therefore increase the likelihood
of early exit and, in this respect alone, tend to reduce inference
latency.

Overall, increasing $s$ reduces the total number of votes while increasing the likelihood of early termination when a pattern is not matched, both of which contribute to lower inference latency. In contrast, increasing $T$ tends to increase the number of votes, which raises inference latency, but it can also increase the likelihood of early termination for non-matching clauses, reducing latency. Consequently, unlike $s$, the effect of $T$ on inference latency is not necessarily monotonic, as these two factors have opposing influences.

\subsection{Training Strategy for Optimal Accuracy-Latency Trade-off} \label{sec:strategy}

\begin{algorithm}[t]
\caption{Latency-aware $(T,s)$ search. Each call to \textsc{Evaluate} trains and
tests one distinct $(T,s)$ (memoized), and counts as one evaluation.}
\label{alg:ts_search}
\begin{algorithmic}[1]
\Require start $(T_0,s_0)$; accuracy threshold $A_{\mathrm{th}}$; steps $\Delta_T,\Delta_s$; upward-probe budget $P$; bound $T_{\min}$
\Ensure feasible $(T^\star,s^\star,a^\star)$ with the largest $s$
\State $best \gets \varnothing$;\ \ $(T,s) \gets (T_0,s_0)$
\While{$T \ge T_{\min}$}
  \State $a \gets \Call{Evaluate}{T,s}$
  \If{$a \ge A_{\mathrm{th}}$}\ $best \gets \Call{Keep}{best,(T,s,a)}$ \EndIf
  \State $climb \gets \varnothing$
  \For{$k \gets 1$ \textbf{to} $P$} \Comment{probe upward to cross accuracy dips}
    \State $s_k \gets s + k\,\Delta_s$
    \State $a_k \gets \Call{Evaluate}{T,s_k}$
    \If{$a_k \ge A_{\mathrm{th}}$}
      \State $best \gets \Call{Keep}{best,(T,s_k,a_k)}$
      \State $climb \gets s_k$ \Comment{keep the highest feasible $s$}
    \EndIf
  \EndFor
  \If{$climb \neq \varnothing$}
    \State $s \gets climb$ \Comment{larger $s$ $\Rightarrow$ earlier exit $\Rightarrow$ lower latency}
  \ElsIf{$a \ge A_{\mathrm{th}}$} 
    \State $T \gets T - \Delta_T$ \Comment{column ceiling; smaller $T$ tolerates larger $s$ (Sec.~\ref{subsec:ts_accuracy})}
  \Else\ \Comment{no feasible cell in this column}
    \State \textbf{break}
  \EndIf
\EndWhile
\State \Return $best$
\end{algorithmic}
\end{algorithm}

Existing $(T,s)$ tuning for TMs targets maximum test accuracy~\cite{tarasyuk2023systematic}.
Latency, however, is governed by a different factor: as shown in
Sec.~\ref{subsec:ts_latency}, a larger $s$ raises the include density of
the clause expression, so clauses evaluate to zero earlier and inference
exits sooner. We therefore treat accuracy as a \emph{constraint} rather
than an objective, and seek the feasible $(T,s)$ combination with the largest $s$.

This is possible because the two objectives are decoupled. Accuracy is
approximately preserved when $s$ is increased and $T$ decreased together
(Section~\ref{subsec:ts_accuracy}), while latency is governed primarily by $s$. The feasible region can hence be traversed towards larger $s$ without
violating the accuracy constraint.

Algorithm~\ref{alg:ts_search} follows this direction. The search is
initialized at a known accuracy-optimal $(T,s)$ combination~\cite{tarasyuk2023systematic},
which is by construction feasible. It then climbs $s$ at the current $T$,
steps $T$ left when no higher $s$ is feasible, and stops when a column has
no feasible cell. Probing $P$ steps upward rather than one lets the climb
cross isolated accuracy dips, where a single infeasible $s$ is followed by
feasible ones above it.

Each distinct $(T,s)$ combination is trained once and counts as one evaluation, so
the number of evaluations equals the number of TM models a reproduction must
train. Exhaustive search must evaluate every combination in the grid, whereas the
directed search evaluates only those along its trajectory.

\section{Experimental Results}

\subsection{Setup}  \label{subsec:set_up}

\subsubsection{Datasets and Booleanization}
The proposed training strategy is validated using six datasets, including Digits~\cite{pen-based_recognition_of_handwritten_digits_81}, MNIST~\cite{Deng2012MNIST}, Kuzushiji-MNIST (KMNIST)~\cite{clanuwat2018deep}, Fashion-MNIST (FMNIST)~\cite{xiao2017fashion}, CIFAR2~\cite{krizhevsky2009learning}, and Keyword Spotting (KWS)~\cite{warden2018speech}. The Booleanization methods are as follows.

For Digits, MNIST, FMNIST, and KMNIST, the raw grayscale pixel values are Booleanized using a fixed threshold, resulting in a 1-bit Boolean representation for each pixel.

For CIFAR2, a binary classification variant of CIFAR10 consisting of animal and vehicle classes, we first extract Histogram of Oriented Gradients (HOG) features. The cell size, block size, and block stride are set to (8$\times$8), (24$\times$24), and (8$\times$8), respectively, producing 36 histograms for each (32$\times$32)-pixel image. Each histogram contains 9 bins, and each bin is Booleanized into a 1-bit feature using two quantiles, resulting in a total of 324 Boolean features.

For KWS, we first perform a short-time Fourier transform (STFT) using a frame length of 0.16 s and a frame step of 0.03 s. From each frame, 13 Mel-Frequency Cepstral Coefficients (MFCCs) are extracted, and each coefficient is encoded using a 3-bit thermometer code. This Booleanization process produces a total of 1,131 Boolean features.

The Booleanized datasets and the corresponding source code for the Booleanization process are openly available at: \href{https://github.com/nsd5g13/ImageBooleanizer}{https://github.com/nsd5g13/ImageBooleanizer}.

\subsubsection{TM Training}

All models are trained using the Python Tsetlin Machine (pyTsetlinMachine) implementation. For each dataset, the hyperparameter combination $(T, s)$ is adopted from prior empirical studies~\cite{tarasyuk2023systematic}, which have been shown to yield near-optimal test accuracy for the corresponding configuration. Adopting these established $(T, s)$ values provides an accuracy-optimized starting point for our subsequent hyperparameter search. The specific $(T, s)$ settings and their corresponding accuracy for each dataset are given in Table ~\ref{tab:acc_latency}.

The dataset and model configurations used in this study are as follows. MNIST, FMNIST, and KMNIST each comprise 10 classes with 784 raw features, trained with 300 clauses per class. Digits also contains 10 classes, but with a much lower feature dimensionality of 64, and is trained with 50 clauses per class. CIFAR2, a binary classification task, comprises 324 features and requires a substantially larger clause budget of 1,000 clauses per class to capture its greater pattern complexity. KWS comprises 8 classes and 377 raw features; quantile binning is applied for Booleanization, converting each feature into a 3-bit thermometer code, and the corresponding model is trained with 500 clauses per class.

In this work, we set the step sizes in Algorithm~1 to $\Delta T = \Delta s = 1$ and the upward-probe budget $P = 2$, so that each column evaluates the
current cell together with two cells above it---three $(T, s)$
combinations in total---before stepping $T$ to the left.

\subsubsection{Deployment and Evaluation}

We employ gem5 v25.0, a cycle-level computer architecture simulator, to conduct our latency evaluation~\cite{10.1145/2024716.2024718}. It is configured in syscall emulation (SE) mode, targeting the ARMv8-A (AArch64) instruction set with the O3CPU model — gem5's timing-accurate, out-of-order CPU model. The bitwise inference kernel is compiled with a 32-bit packed representation, so each clause is evaluated in $\lceil n/32 \rceil$ word-level iterations. The cache hierarchy consists of 64 kB, 4-way set-associative L1 instruction and data caches with a 64-byte cache line size, a configuration typical of application-class cores used in edge deployments~\cite{francis2018raspberry, aalishah2025edgenavmamba}.

We use the number of simulated clock cycles reported by gem5 as our latency metric. Since this cycle count is deterministic and unaffected by external randomness, it provides a consistent and repeatable basis for comparison across different $(T, s)$ configurations. All experiments are therefore conducted within this simulation environment rather than on a physical device. This choice is made to avoid the non-determinism inherent to real hardware, where OS-level background processes and scheduling overhead constitute a fixed but hard-to-isolate portion of the measured runtime, potentially confounding the latency attributable to the model configuration itself.

\subsection{Design Space Exploration}
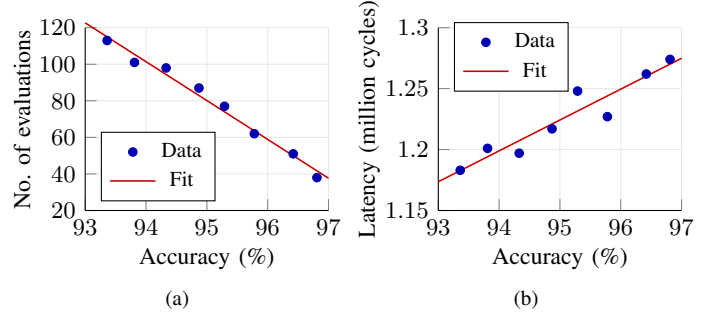
\begin{figure}[t]
  \centering

  \subfloat[
  ]{%
    \begin{tikzpicture}
      \begin{axis}[
        xlabel={Accuracy (\%)},
        ylabel={No. of evaluations},
        xmin=93, xmax=97, xtick={93,94,95,96,97},
        ymin=20, ymax=120, ytick={20,40,60,80,100,120},
        axis x line*=bottom, axis y line*=left,
        tick label style={font=\small},
        label style={font=\small},
        grid=both, grid style={gray!18},
        height=4cm,
        width=4.8cm,
        ylabel style={yshift=-2.5ex},
        xlabel style={yshift=1ex},
        clip=false,
legend style={
at={(0.3,0.05)},anchor=south,font=\footnotesize
       }
      ]
        \addplot[only marks, mark=*, mark size=1.6pt, blue!70!black]
          coordinates {
            (93.36,113) (93.81,101) (94.33,98) (94.87,87)
            (95.29,77)  (95.78,62)  (96.42,51) (96.81,38)
          };
          \addlegendentry{Data}
        \addplot[red!80!black, semithick, domain=93:97, samples=2]
          {-21.247*x + 2098.615};
          \addlegendentry{Fit}
      \end{axis}
    \end{tikzpicture}%
  }
  \subfloat[
  ]{%
    \begin{tikzpicture}
      \begin{axis}[
        xlabel={Accuracy (\%)},
        ylabel={Latency (million cycles)},
        xmin=93, xmax=97, xtick={93,94,95,96,97},
        ymin=1.15, ymax=1.30, ytick={1.15,1.20,1.25,1.30},
        yticklabel style={/pgf/number format/fixed,
                          /pgf/number format/precision=2},
        axis x line*=bottom, axis y line*=left,
        tick label style={font=\small},
        label style={font=\small},
        grid=both, grid style={gray!18},
        height=4cm,
        width=4.8cm,
        ylabel style={yshift=-2ex},
        xlabel style={yshift=1ex},
        legend style={
        at={(0.3,0.65)}, anchor=south,
        font=\footnotesize
        }
      ]
        \addplot[only marks, mark=*, mark size=1.6pt, blue!70!black]
          coordinates {
            (93.36,1.183) (93.81,1.201) (94.33,1.197) (94.87,1.217)
            (95.29,1.248) (95.78,1.227) (96.42,1.262) (96.81,1.274)
          };
        \addlegendentry{Data}
        \addplot[red!80!black, semithick, domain=93:97, samples=2]
          {0.02533*x - 1.18209};
        \addlegendentry{Fit}
      \end{axis}
    \end{tikzpicture}%
  }

  \caption{
    (a) Number of evaluations and (b) latency under varying accuracy constraints. Relaxing the accuracy constraint reduces inference latency while increasing the search effort. Both trends exhibit an approximately linear relationship, with $R^2$ values of 0.98 and 0.90, respectively.}
  \label{fig:dse}
\end{figure}

Fig.~\ref{fig:dse} sweeps the accuracy constraint on MNIST and
records, for each target, the number of \((T,s)\) configurations the
search evaluates~(a) and the latency of the returned configuration~(b).
Both quantities vary approximately linearly with the constraint
(\(R^2 = 0.98\) and \(0.90\)), but in opposite directions: tightening
the target from 93.36\% to 96.81\% reduces the evaluation count from
113 to 38 while raising latency from 1.183 to 1.274 million cycles.
The opposing trends expose a trade-off. Relaxing the constraint widens
the feasible band, allowing the search to advance further along the
\(s\)-axis before the accuracy floor is reached; the resulting
higher-\(s\) configuration exits earlier and therefore runs faster, but reaching it requires evaluating more intermediate points, increasing the search effort. A designer can thus exchange a small
accuracy margin for lower latency at a predictable search cost. Because
both relationships are approximately linear, the search cost and
latency of an arbitrary target within this range can be estimated by
interpolation, without executing the full sweep.

\subsection{Accuracy--Latency Trade-off} 

\definecolor{cenergy}{HTML}{C6E0B4} 

\begin{table*}[]
\centering
\caption{Accuracy--Latency trade-off across datasets under different accuracy-drop budgets}
\label{tab:acc_latency}
\footnotesize

\setlength{\tabcolsep}{4pt}

\begin{tabular}{l||ccc||cccccccccc} 
\hline
\multicolumn{1}{c||}{\multirow{3}{*}{Dataset}} & \multicolumn{3}{c||}{\multirow{2}{*}{Baseline}} & \multicolumn{10}{c}{This Work}                                                                                                                      \\ \cline{5-14} 
\multicolumn{1}{c||}{}                         & \multicolumn{3}{c||}{}                          & \multicolumn{5}{c||}{Max. acc. drop: 0.5\%}                                                          & \multicolumn{5}{c}{Max. acc. drop: 2\%}                                       \\ \cline{2-14} 
\multicolumn{1}{c||}{}                         & $(T, s)$         & \makecell{Acc.\\(\%)}         & \makecell{Latency\\(kilo cycles)}        & \makecell{No. of\\evals} & $(T, s)$ & \makecell{Acc.\\(\%)} & \makecell{Latency\\(kilo cycles)} & \multicolumn{1}{c||}{\makecell{Latency\\reduct.(\%)}} & \makecell{No. of\\evals} & $(T, s)$ & \makecell{Acc.\\(\%)} & \makecell{Latency\\(kilo cycles)} & \makecell{Latency\\reduct.(\%)} \\ \hline
\rowcolor{gray!15} MNIST                                 &      (14,11.10)       &     97.26         &         1348.27        &        38         &    (8,25.10)   &   96.81   &     1273.90    & \multicolumn{1}{c||}{5.52}                    &        77         &   (4,52.10)    &   95.29   &    1248.23     &            7.42         \\
KMNIST                                         &     (14,11.10)          &      85.97        &         1305.00        &        24         &   (11,20.10)    &   85.53   &   1220.68      & \multicolumn{1}{c||}{6.46}                    &       38          &   (9,28.10)   &   84.07   &    1171.33     &         10.24            \\
\rowcolor{gray!15} FMNIST                                         &       (14,11.10)        &      85.97        &       1147.96          &       17          &   (13,19.10)    &   85.48  &   1072.36      & \multicolumn{1}{c||}{6.59}                    &       88          &   (5,66.10)    &  84.25    &    1010.07     &         12.01            \\
Digits                                         &       (7, 6.56)        &      95.28        &        94.18         &         10        &     (6,7.56)  &  95.00    &    93.94     & \multicolumn{1}{c||}{0.25}                    &        9         &    (7,9.56)   &   93.33   &   91.73      &          2.60           \\
\rowcolor{gray!15} CIFAR2                                        &       (24,14.15)         &       89.24       &        438.38         &       63          &   (20,59.15)    &   88.75   &     391.40    & \multicolumn{1}{c||}{10.72}                    &       100          &   (18,95.15)    &   87.52   &     366.74    &          \cellcolor{cenergy} \textbf{16.34}          \\
KWS                                            &        (18,12.39)       &    \textbf{70.56}        &       1892.98          &      17           &   (17,20.39)    &   \cellcolor{cenergy} \textbf{70.50}   &     1766.71    & \multicolumn{1}{c||}{6.67}                    &        26         &    (16,26.39)   &   68.94   &    1689.00     &          10.78           \\ \hline
\end{tabular}
\end{table*}

Table~\ref{tab:acc_latency} reports, for each dataset, the accuracy-optimal baseline configuration together with the configurations returned by the proposed search under two accuracy-drop budgets, 0.5\% and 2\%. Each entry lists the \((T,s)\) combination, the resulting test accuracy, and the inference latency in kilo-cycles obtained from the gem5 model of Section \ref{subsec:set_up}. The baseline is the peak-accuracy configuration; the two budgeted columns instead treat this accuracy as a constraint and report the largest $s$ that keeps the accuracy drop within the stated budget, alongside the number of \((T,s)\) evaluations the search consumed and the latency reduction relative to the baseline.

Across all six datasets, the returned configurations increase $s$ relative to the baseline and, in most cases, also reduce $T$, placing them to the upper-left of the baseline in the $(T,s)$ plane. Since latency is governed primarily by $s$, this movement translates directly into lower inference latency. Under the 0.5\% budget, the reduction reaches 10.72\% on CIFAR2, while on Digits it is only 0.25\%; relaxing the budget to 2\% increases the reduction on every dataset, with CIFAR2 again leading at 16.34\% and Digits lowest at 2.60\%. The consistently small reduction on Digits reflects its low feature dimension, which leaves little room to increase $s$ before the accuracy constraint is violated. The larger admissible accuracy drop simply permits a larger $s$, and hence a shorter expected inference path, at the cost of the additional accuracy conceded to the budget. KWS is a clear instance of the intended behavior: under the 0.5\% budget its accuracy is essentially unchanged (70.56\% to 70.50\%), yet increasing $s$ from 12.39 to 20.39 lowers latency by 6.67\%, so a lower-latency configuration is obtained with virtually no accuracy penalty.

The search reaches these configurations at a small cost, between 9 and 100 evaluations, far below an exhaustive sweep of the $(T,s)$ grid. As expected, the looser 2\% budget generally admits a longer trajectory and therefore more evaluations, since the search can climb further in $s$ before the accuracy constraint is violated. To bound this cost, each search is capped at 100 evaluations; CIFAR2 under the 2\% budget reaches this cap, so its reported configuration is the best found within the budget rather than a guaranteed largest-$s$ solution.

\section {Conclusions and Discussions} \label{sec:conc}

In this work, we systematically evaluated how the two TM training
hyperparameters $T$ and $s$ jointly affect classification accuracy
and inference latency on a bitwise, early-exit software implementation.
Based on this, we showed that adjusting $T$ and $s$ along their
inversely proportional relationship preserves
the accuracy of an accuracy-optimal configuration. Inference
latency, however, is governed separately: it is lowered primarily
by increasing $s$, which raises the early-exit probability.
Exploiting this decoupling, we proposed a directed $(T, s)$ search
that treats accuracy as a constraint and advances towards larger
$s$ within a fixed budget of training evaluations. Across six
datasets, with latency measured on a gem5-simulated ARM processor,
the method reduces inference latency by up to 16.34\%
with some datasets showing negligible accuracy loss, while requiring far
fewer configuration evaluations than an exhaustive grid search. As
the search is budget-bounded, its result is a practical low-latency
configuration rather than a guaranteed global optimum.

For other implementations, such as REDRESS, where only included features are encoded in the model representation, the trend observed in Fig.~\ref{fig:Ts_include} also provides guidance for accelerating inference. For a fixed value of $T$, decreasing $s$ reduces the total number of included features, resulting in less information to encode, smaller model storage, and lower inference latency. Therefore, the strategy presented in Section~\ref{sec:strategy} is also applicable to REDRESS, though in the opposite direction, as inference latency can be reduced by decreasing $s$ while adjusting $T$ to maintain accuracy with minimal degradation.

\section*{Acknowledgement}
This work was supported by the Engineering and Physical Sciences Research Council (EPSRC) under Grant EP/X039943/1 (ESTEEM) and Grant EP/X036006/1 (SONNETS).

\bibliographystyle{IEEEtran}
\bibliography{references}

\end{document}